\documentclass[11pt,a4paper]{article}
\usepackage[hyperref]{style/acl2021}
\usepackage{times}
\usepackage{latexsym}

\usepackage{microtype}

\aclfinalcopy 
\def\aclpaperid{2041} 

\usepackage{amsmath}
\usepackage{bm}
\usepackage{booktabs}
\usepackage{float}
\usepackage{graphicx}
\usepackage{subcaption}
\usepackage{tabularx}

\newcommand{\abr}[1]{\textsc{#1}}
\newcommand{\vect}[1]{\bm{\mathbf{#1}}}
\newcommand{\g}{\, | \,}
\newcommand{\flag}[1]{{\setlength{\fboxsep}{0pt}\fbox{\includegraphics[height=0.24cm,width=0.36cm]{figures/flags/#1.pdf}}}}
\DeclareMathOperator*{\argmax}{arg\,max}

\usepackage{colortbl}
\definecolor{lightgrey}{rgb}{0.85,0.85,0.85}

\newcommand{\langnum}{ten} 
\newcommand{\name}[0]{\abr{mrs}} 

\title{A Dataset and Baselines for Multilingual Reply Suggestion}

\author{
Mozhi Zhang\thanks{\hspace{1.5mm}Work mostly done as an intern at Microsoft Research.} \\
University of Maryland \\
\small\tt mozhi@cs.umd.edu \And
Wei Wang\thanks{\hspace{1.5mm}Work done at Microsoft Research.} \\
Qualtrics \\
\small\tt wwang@qualtrics.com \And
Budhaditya Deb \\
Microsoft AI \\
\small\tt budeb@microsoft.com \AND
Guoqing Zheng \\
Microsoft Research \\
\small\tt zheng@microsoft.com \And
Milad Shokouhi \\
Microsoft AI \\
\small\tt milads@microsoft.com \And
Ahmed Hassan Awadallah \\
Microsoft Research \\
\small\tt hassanam@microsoft.com
}

\date{}

\begin{document}
\maketitle
\begin{abstract}
Reply suggestion models help users process emails and chats faster.
Previous work only studies English reply suggestion.
Instead, we present \name{}, a multilingual reply suggestion dataset with \langnum{} languages.
\name{} can be used to compare two families of models: 
1) retrieval models that select the reply from a fixed set and 2) generation models that produce the reply from scratch.
Therefore, \name{} complements existing cross-lingual generalization benchmarks that focus on classification and sequence labeling tasks.
We build a generation model and a retrieval model as baselines for \name{}.
The two models have different strengths in the monolingual setting, and they require different strategies to generalize across languages.
\name{} is publicly available at \url{https://github.com/zhangmozhi/mrs}.
\end{abstract}

\section{Multilingual Reply Suggestion}
\label{sec:intro}

\begin{figure}[t]
    \centering
    \includegraphics[width=\linewidth]{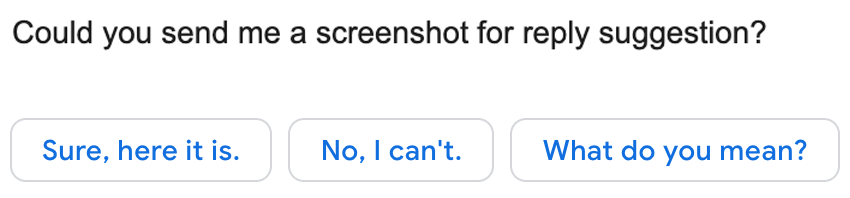}
    \caption{An example of reply suggestion system.  User can click on the suggestions for a quick reply.}
    \label{fig:reply}
\end{figure}

Automated reply suggestion is a useful feature for email and chat applications.
Given an input message, the system suggests several replies, and users may click on them to save typing time (Figure~\ref{fig:reply}).
This feature is available in many applications including Gmail, Outlook, LinkedIn, Facebook Messenger, Microsoft Teams, and Uber.

Reply suggestion is related to but different from open-domain dialog systems or chatbots~\citep{adiwardana2020humanlike,huang2020challenges}.
While both are conversational \abr{ai} tasks~\citep{gao2019neural}, the goals are different:
reply suggestion systems help the user quickly reply to a message,
while chatbots aim to \emph{continue} the conversation and focus more on multi-turn dialogues.

Ideally, we want our model to generate replies in any language.
However, reply suggestion models require large training sets, so previous work mostly focuses on English~\citep{kannan2016smart,henderson2017efficient,deb2019diversifying}.
To investigate reply suggestion for other languages with possibly limited data,
we build a multilingual dataset, dubbed \name{} (\textbf{M}ultilingual \textbf{R}eply \textbf{S}uggestion).
From publicly available Reddit threads, we extract message-reply pairs, response sets, and machine-translated examples in \langnum{} languages~(Table~\ref{tab:lang}).

One interesting aspect of the reply suggestion problem is that there are two modeling approaches.
Some models follow the retrieval framework and select the reply from a predetermined response set~\citep{henderson2017efficient}.
Others follow the generation framework and generate the reply from scratch~\citep{kannan2016smart}.
The two approaches have different advantages.
Generation models are more powerful because they are not constrained by the response set.
In comparison, retrieval models are easier to train and runs faster, and a curated response set guarantees the coherence and the safety of the model output.

\begin{table*}[t]
    \centering
    \begin{tabular}{lllrrr}
        \toprule
         Language & Code & Family & Examples & Tokens & Response Set\\ \midrule
         \flag{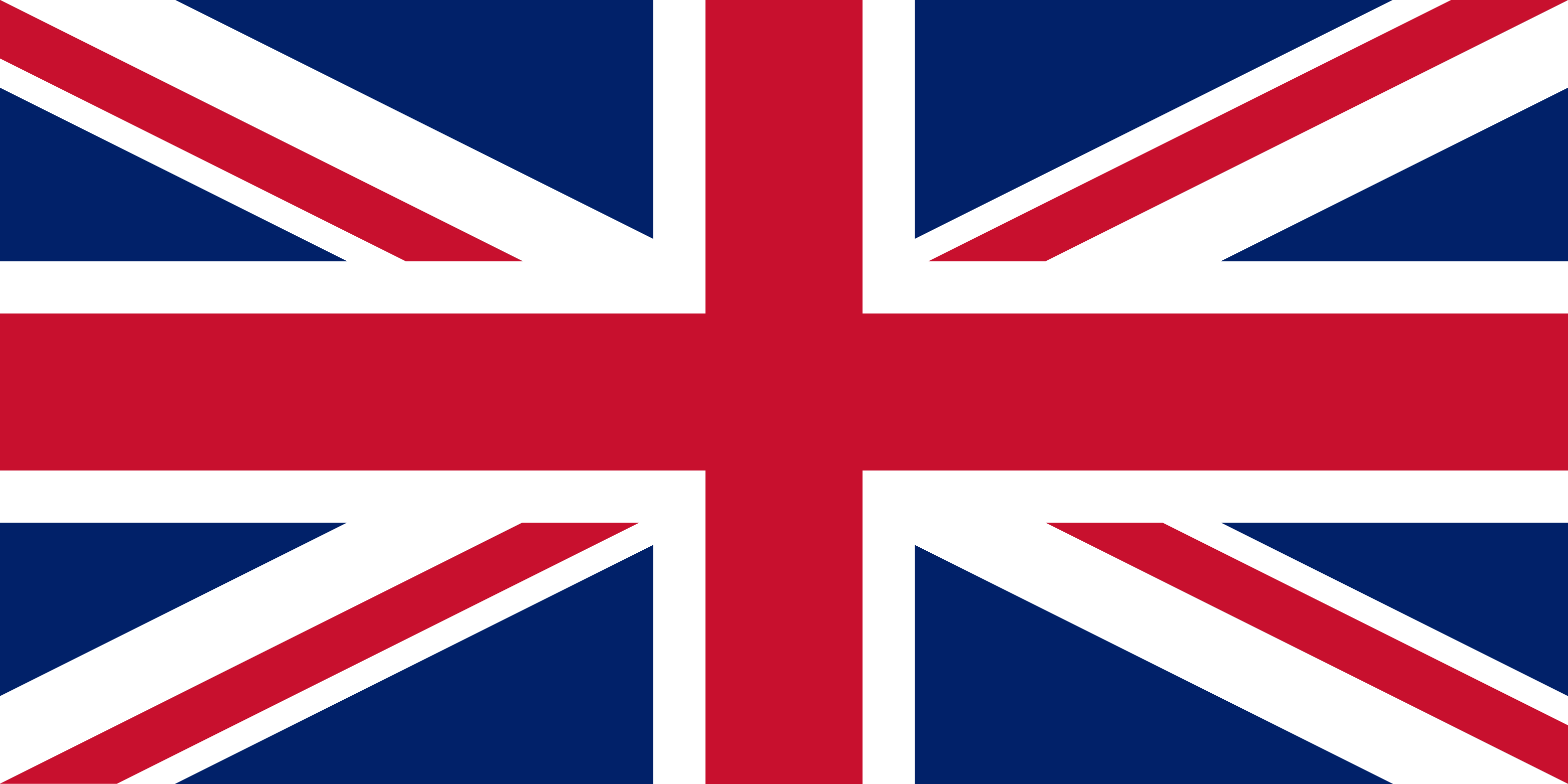} English & \abr{en} & West Germanic & 48,750,948 & 1,700,066,696 & 36,997\\
         \flag{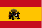} Spanish & \abr{es} & Romance & 2,325,877 & 195,424,517 & 45,152\\
         \flag{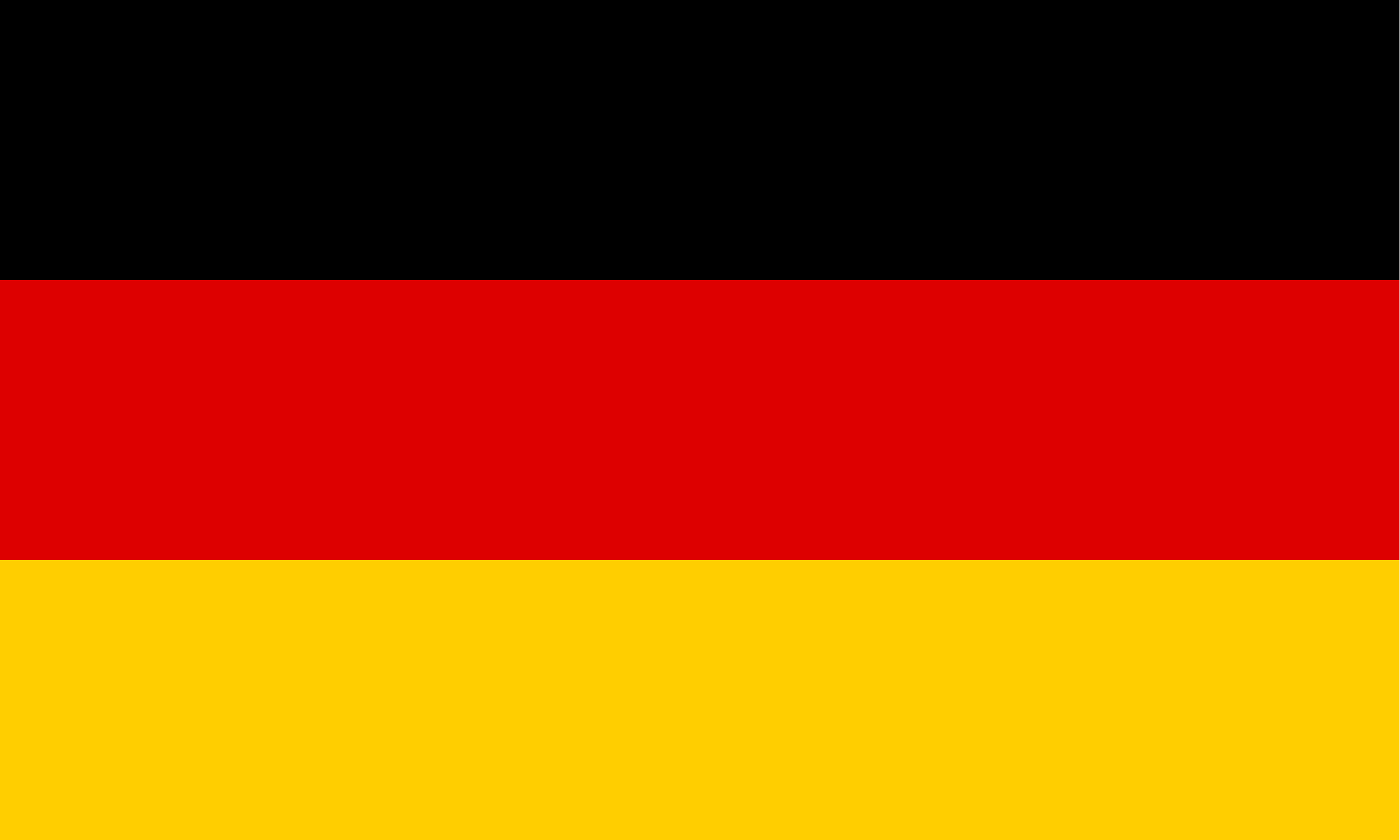} German & \abr{de} & West Germanic & 1,864,688 & 118,711,662 & 34,747\\
         \flag{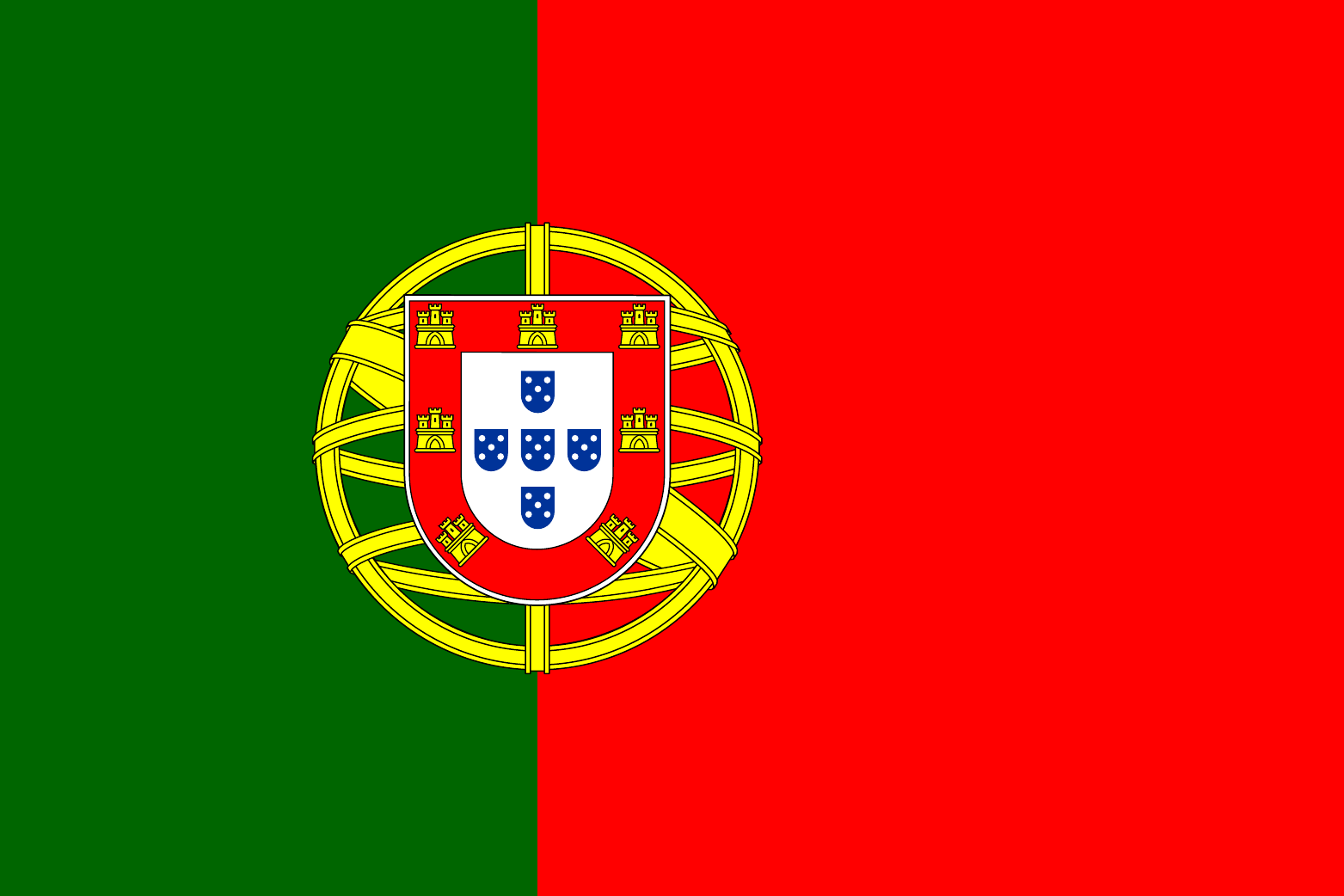} Portuguese & \abr{pt} & Romance & 1,822,594 & 114,642,809 & 45,225\\
         \flag{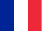} French & \abr{fr} & Romance & 1,396,806 & 133,068,740 & 32,350\\
         \flag{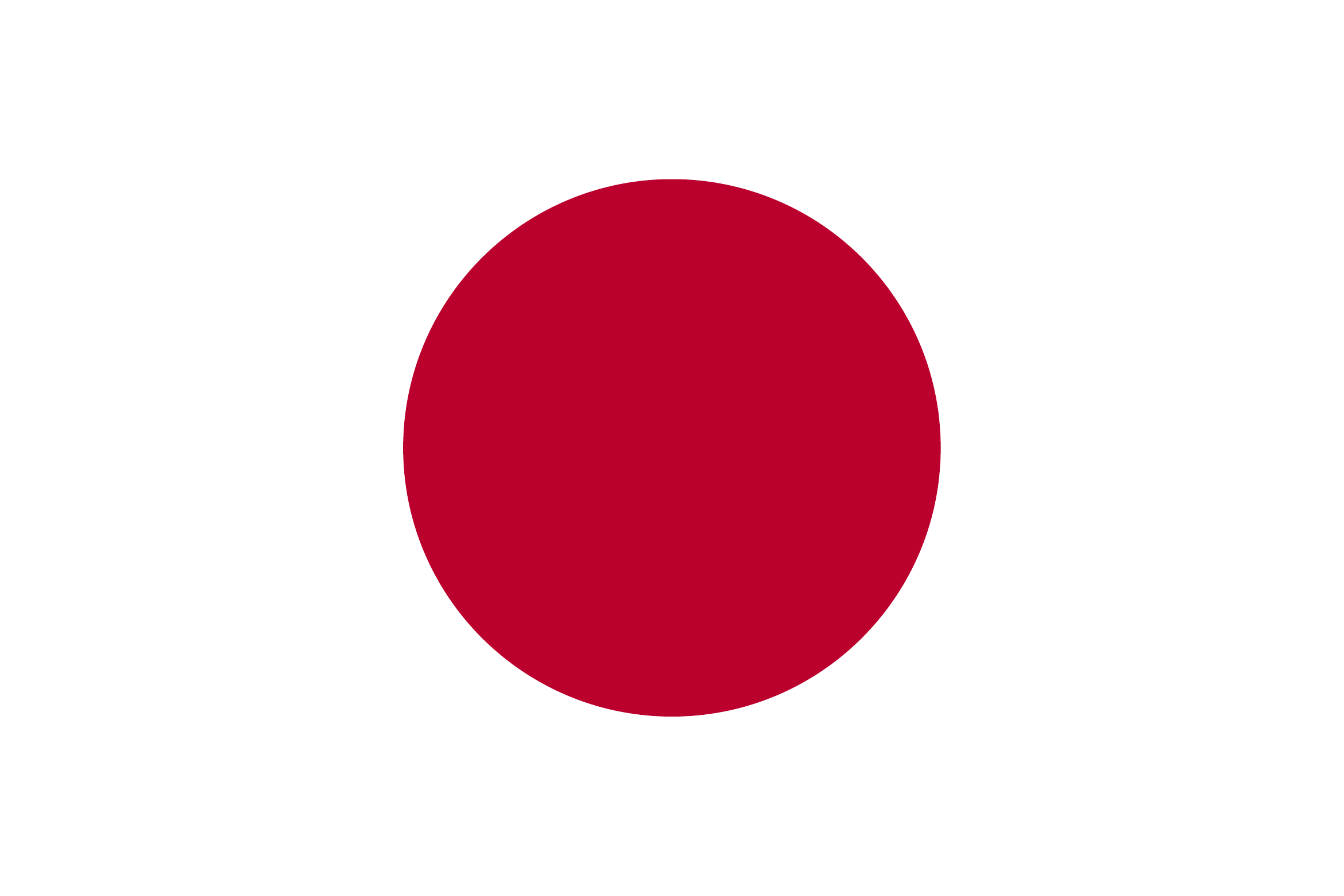} Japanese & \abr{ja} & Japonic & 727,668 & 46,124,966 & 38,817\\
         \flag{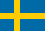} Swedish & \abr{sv} & North Germanic & 738,254 & 47,845,497 & 32,165\\
         \flag{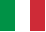} Italian & \abr{it} & Romance & 736,296 & 58,715,043 & 31,855\\
         \flag{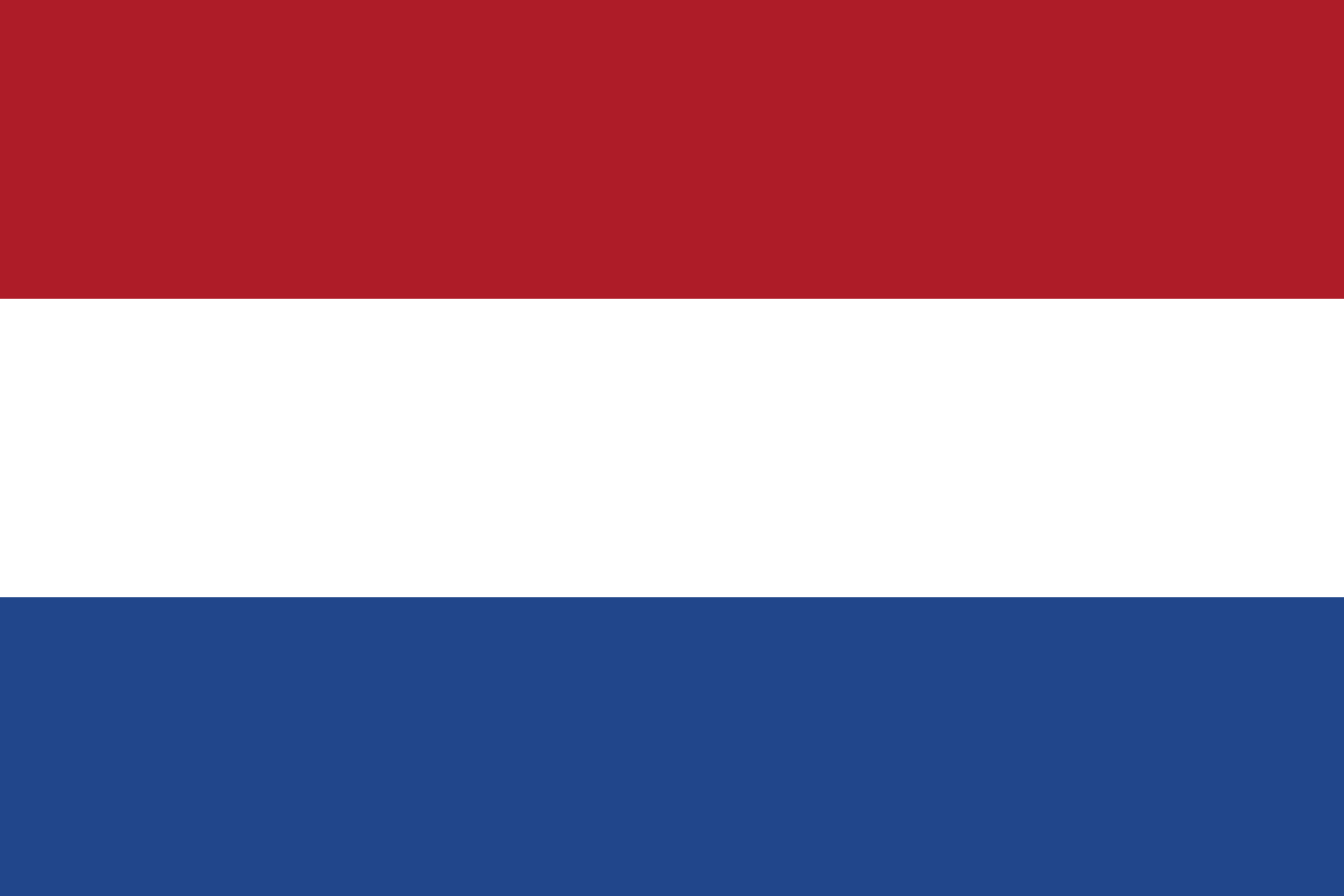} Dutch & \abr{nl} & West Germanic & 638,634 & 43,847,547 & 32,293\\
         \flag{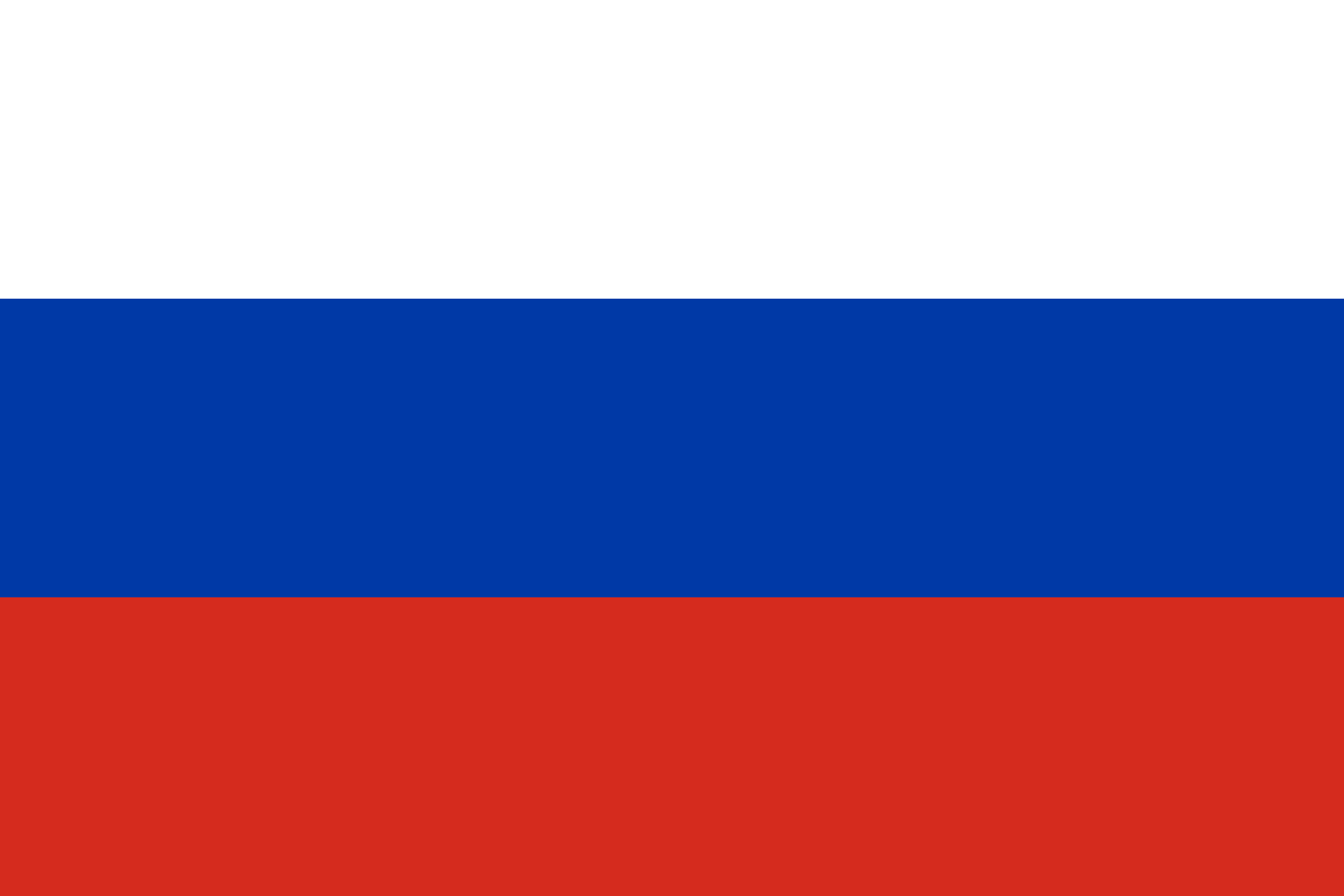} Russian & \abr{ru} & East Slavic & 516,739 & 23,109,295 & 31,475\\
         \bottomrule
    \end{tabular}
    \caption{Dataset statistics for \name{}.
    We collect Reddit message-reply pairs for \langnum{} language.
    For each language, we use 80\% examples for training, 10\% for validation, and 10\% for testing.
    We then create response sets for retrieval models.
    We also use \abr{mt} to translate nineteen million English training examples to other languages.
    }
    \label{tab:lang}
\end{table*}

The two frameworks make reply suggestion an interesting task for studying cross-lingual generalization.
Most cross-lingual generalization benchmarks use classification and sequence labeling tasks~\citep{tjong2002introduction,nivre2016universal,strassel2016lorelei,conneau2018xnli,li2018mldoc,clark2020tydi,hu2020xtreme,lewis2020mlqa}.
In contrast, reply suggestion has two formulations that require different cross-lingual generalization strategies.
While some recent work explores cross-lingual transfer learning in generation tasks, 
the tasks are \emph{extractive}; i.e., the output often has significant overlap with the input.
These tasks include news title generation, text summarization, and question generation~\citep{chi2020cross,liang2020xglue,scialom2020mlsum}.
Reply suggestion is more challenging because the reply often does not overlap with the message (Figure~\ref{fig:reply}), so the model needs to address different cross-lingual generalization challenges (Section~\ref{ssec:xling_exp}).

We build two baselines for \name{}: a retrieval model and a generation model.
We first compare the models in English, where we have abundant training data and human referees.
We evaluate the models with both automatic metrics and human judgments.
The two models have different strengths.
The generation model has higher word overlap scores and is favored by humans on average, but inference is slower, and the output is sometimes contradictory or repetitive~\citep{holtzman2020curious}.
In contrast, the retrieval model is faster and always produces coherent replies, but the replies are sometimes too generic or irrelevant due to the fixed response set.

Next, we test models in other languages.
We compare different training settings and investigate two cross-lingual generalization methods: initializing with pre-trained multilingual models~\citep{wu2019beto,conneau2020unsupervised,liang2020xglue} and training on machine-translated data~\citep{banea2008multilingual}.
Interestingly, the two models prefer different methods: multilingual pre-training works better for the retrieval model, while the generation model prefers machine translation.

In summary, we present \name{}, a multilingual reply suggestion dataset.
We use \name{} to provide the first systematic comparison between generation and retrieval models for reply suggestion in both monolingual and multilingual settings. 
\name{} is also a useful benchmark for future research in reply suggestion and cross-lingual generalization.

The rest of the paper is organized as follows.
Section~\ref{sec:data} describes the data collection process for \name{}.
Section~\ref{sec:setting} introduces task formulations, experiment settings, and evaluation metrics.
Section~\ref{sec:model} describes the baseline generation and retrieval models.
Section~\ref{sec:result} presents our experiment results.
Section~\ref{sec:future} discusses how \name{} can help future research.
\section{Dataset Construction}
\label{sec:data}

To study reply suggestion in multiple languages, we build \name{}, a dataset with message-reply pairs based on Reddit comments.
The dataset is available at \url{https://github.com/zhangmozhi/mrs}.

We download Reddit comments between January 2010 and December 2019 from the Pushshift Reddit dataset~\citep{baumgartner2020pushshift}.\footnote{\url{https://files.pushshift.io/reddit/comments}}
We extract message-reply pairs from each thread by considering the parent comment as an input message and the response to the comment as the reference reply.
We remove comments starting with \emph{[removed]} or \emph{[deleted]}, which are deleted messages.
We also skip comments with a rating of less than one, since they are likely to contain inappropriate content.

After extracting examples, we identify their languages with fastText language detector~\citep{joulin2016fasttext}.
For each example, we run the model on the concatenation of the message and the reply.
We discard low-confidence examples where none of the languages has a score higher than 0.7.
For the remaining examples, we use the highest-scoring label as the language.

We only use English data from 2018 because English data is abundant on Reddit.
Non-English examples are much more scarce, so we use data from the last ten years.
We select the top \langnum{} languages with at least 100K examples.
We create three splits for each language: 80\% examples for training, 10\% for validation, and 10\% for testing.

Table~\ref{tab:lang} shows some dataset statistics.
\name{} is heavily biased towards English.
We have more than 48 million English examples, but fewer than one million examples for half of the languages.
This gap reflects a practical challenge for reply suggestion---we do not have enough data for most languages in the world.
Nevertheless, we can use \name{} to test models in different multilingual settings, including cross-lingual transfer learning, where we build non-English reply suggestion models from English data (Section~\ref{ssec:setting}).

We also build response sets and filter out toxic examples.  We describe these steps next.

\subsection{Response Set}
\label{ssec:response}

We build a response set of 30K to 50K most frequent replies for each language, which are used in the retrieval model.
We want the response set to cover generic responses, so we select replies that appear at least twenty times in the dataset.
This simple criterion works well for English, but the set is too small for other languages.
For non-English languages, we augment the response set by translating the English response set to other languages with Microsoft Translator.
The non-English response set is sometimes smaller than the English set, because different English responses may have the same translation.

\subsection{Filtering Toxic Examples}
\label{ssec:filter}

Exchanges on Reddit are sometimes uncivil, inappropriate, or even abusive~\citep{massanari2017gamergate,mohan2017impact}.
We try to filter out toxic contents, as they are not desirable for reply suggestion systems.

We use two toxicity detection models.
First, we use an in-house multilingual model.
The model is initialized with multilingual \abr{bert}~\citep[\abr{mbert}]{devlin2019bert} and fine-tuned on a mixture of proprietary and public datasets with toxic and offensive language labels.
The model outputs a score from zero to one, with a higher score corresponding to a higher level of toxicity.
Second, we use Perspective \abr{api}\footnote{\url{https://www.perspectiveapi.com}}, a publicly available model. 
Perspective \abr{api} has limited free access (one query per second), so we only use the \abr{api} on the English validation, test, and response set.
For other languages, we rely on our in-house model.
We filter message-reply pairs if it has greater than 0.9 score according to the in-house model, or greater than 0.5 score according to Perspective \abr{api}~\citep{gehman2020real}.
About one percent of examples are filtered.
After filtering the data, we manually validate three hundred random examples and do not find any toxic examples, which confirms that our filter method have a high recall.

While we hope the filtered dataset leads to better reply suggestion models, existing filtering methods are not perfect and can introduce other biases~\citep{dixon2018measuring,sap2019risk,hutchinson2020social}.
Therefore, models trained on all \name{} data may still have undesirable behavior.
\name{} is intended to be used as a benchmark for testing cross-lingual generalization of generation and retrieval models.
\textbf{The dataset should not be directly used in production systems.}
To use the dataset in practice, additional work is required to address other possible biases and toxic or inappropriate content that may exist in the data.
\section{Experiment Settings}
\label{sec:setting}

After presenting the dataset, we explain how we use \name{} to compare reply suggestion models.
We describe the two frameworks for reply suggestion, our experiment settings, and evaluation metrics.

\subsection{Task Formulation}
\label{ssec:framework}

In reply suggestion, the input is a message $\vect{x}$, and the output is one or more suggested replies $\vect{y}$.
In practice, reply suggestion systems can choose to not suggest any replies.
This decision is usually made by a separate trigger model~\citep{kannan2016smart}.
In this paper, we focus on reply generation, so we assume that the models always need to suggest a fixed number of replies.
Reply suggestion can be formulated as either a \emph{retrieval} problem or a \emph{generation} problem.

\paragraph{Retrieval Model.}
A retrieval model selects the reply $\vect{y}$ from a fixed response set~$\mathcal{Y}$ (Section~\ref{ssec:response}).
Given an input message $\vect{x}$, the model computes a relevance score $\vect{\Theta}_{\vect{x}\vect{y}}$ for each candidate reply $\vect{y}\in\mathcal{Y}$.
The model then selects the highest-scoring replies as suggestions; e.g., the top-1 reply is $\argmax_{\vect{y} \in \mathcal{Y}} \Theta_{\vect{x}\vect{y}}$.

\paragraph{Generation Model.}
A generation model generates the reply $\vect{y}$ from scratch.
Generation models usually follow the sequence-to-sequence framework~\citep[\abr{seq2seq}]{sutskever2014sequence}, which generates $\vect{y}$ token by token.
Given an input message $\vect{x} = (x_1, x_2, \cdots, x_n)$ of $n$ tokens, a \abr{seq2seq} model estimates the probability of a reply $\vect{y} = (y_1, y_2, \cdots, y_m)$ of $m$ tokens as following:
\begin{equation}
\label{eq:likelihood}
p(\vect{y} \g \vect{x}) = \prod_{i=1}^{m} p(y_i \g \vect{x}, y_{<i}).
\end{equation}
The model computes probability for the next token $p(y_i \g \vect{x}, y_{<i})$ based on the input $\vect{x}$ and the first~$(i-1)$ tokens of the output $\vect{y}$.
The model is trained to maximize the probability of reference replies in the training set.
At test time, we find the top replies that approximately maximize \eqref{eq:likelihood} with beam search.

The two models have different strengths.
The generation model is more flexible, but the retrieval model is faster~\citep{henderson2017efficient}, and the output can be controlled by curating the response set~\citep{kannan2016smart}.

We compare a retrieval model and a generation model as baselines for \name{}.
To our knowledge, we are the first to systematically compare the two models in both monolingual and multilingual settings.  We explain our training settings and metrics next.

\subsection{Training Settings}
\label{ssec:setting}

For each language in \name{}, we train and compare models in four settings.
Future work can experiment with other settings (discussed in Section~\ref{sec:future}).

\paragraph{Monolingual.}
Here, we simply train and test models in a single language.
This setting simulates the scenario where we have adequate training data for the target language. Previous reply suggestion models were only studied in the English monolingual setting.

\paragraph{Zero-Shot.}
Next, we train models in a zero-shot cross-lingual setting.
We train the model on the English training set and use the model on the test set for another language.
This setting simulates the scenario where we want to build models for a low-resource language using our large English set.
To generalize across languages, we initialize the models with pre-trained multilingual models (details in Section~\ref{sec:model}).
These models work well in other tasks~\citep{wu2019beto,liang2020xglue}.
We test if they also work for reply suggestion, as different tasks often prefer different multilingual representations~\citep{zhang2020overfit}.

\paragraph{Machine Translation (\abr{mt}).}
Another strategy for cross-lingual generalization is to train on machine-translated data~\citep{banea2008multilingual}.
We train models on nineteen million English training examples machine-translated to the target language with Microsoft Translator.
We compare against the zero-shot setting to compare the two cross-lingual generalization strategies.

\paragraph{Multilingual.}
Finally, we build a multilingual model by jointly training on the five languages with the most training data: English, Spanish, German, Portuguese, and French.
We oversample non-English training data to have the same number of training examples data across all languages~\citep{johnson2017google}.
We make two comparisons:
1) for the five training languages, we compare against the \emph{monolingual} setting to test whether fitting multiple languages in a single model hurts performance;
and 2) for other languages, we compare against the \emph{zero-shot} setting to check if adding more training languages helps cross-lingual generalization.

\subsection{Evaluation Metrics}
\label{ssec:metric}

The goal of reply suggestion is to save user typing time, so the ideal metrics are click-through rate (\abr{ctr}), how often the user chooses a suggested reply, and time reduction, how much time is saved by clicking the suggestion instead of typing.
However, these metrics require deploying the model to test on real users, which is not feasible at full-scale while writing this paper. Instead, we focus on automated offline metrics that can guide research and model development before deploying production systems. Specifically, we evaluate models using a test set of message-reply pairs.

To identify a good metric, we compare several metrics in a pilot study by deploying an English system.
We collect millions of user interactions and measure Pearson's correlation between \abr{ctr} and automated offline metrics.
The next paragraph lists the metrics.
Based on the study, we recommend weighted \abr{rouge} F1 ensemble (\textbf{\abr{rouge}} in tables), which has the highest correlation with \abr{ctr}.

For the retrieval model, we follow previous work and consider mean reciprocal rank \citep[\abr{mrr}]{kannan2016smart} and precision at one~\citep{henderson2017efficient}.
These metrics test if the model can retrieve the reference response from a random set of responses. Alternatively, we compute \abr{mrr} and precision on a subset of examples where the reference reply is in the response set so that we can directly measure the rank of the reference response in the response set. This set also allows us to compute \abr{mrr} for individual responses, so we can compute macro-\abr{mrr}, the average \abr{mrr} over each response in the set. Higher macro-\abr{mrr} can indicate diversity but has a worse correlation than computing \abr{mrr} over the entire test set. 
For the generation model, we consider model perplexity~\citep{adiwardana2020humanlike}.
Finally, we consider two word overlap scores, \abr{bleu}~\citep{papineni2002bleu} and \abr{rouge}~\citep{lin2004rouge}, which can be used for both retrieval and generation models.

Our pilot study shows that \abr{rouge} has the best correlation.
However, individual \abr{rouge} F1 scores (\abr{rouge-1/2/3}) are sensitive to small changes in sequence lengths (more so because our responses are generally short). Therefore, we use a weighted average of the three scores:
\begin{equation}
\label{eq:rouge}
     \frac{\abr{Rouge-1}}{6} + \frac{\abr{Rouge-2}}{3} + \frac{\abr{Rouge-3}}{2}.
\end{equation}
This weighted score leads to the highest correlation with \abr{ctr}.
Intuitively, the weights balance the differences in the average magnitude of each metric and thus reduce variance on short responses.

Popular reply suggestion systems (such as Gmail and Outlook) suggest three replies for each message, while the user only selects one. To simulate this setting, we predict three replies for each message.
For the retrieval model, we use the three highest-scoring replies from the response set.
For the generation model, we use top-three results from beam search.
Out of the three replies, we only use the reply with the highest \abr{rouge} compared to the reference reply when computing the final metrics; i.e., the model only has to provide one ``correct'' reply to have a full score.

We compare models primarily with \abr{rouge}, since the metric has the best correlation in the pilot study.
Nevertheless, word overlap scores have known limitations~\citep{liu2016evaluate}, as there are different ways to reply to a message.
We encourage future research to investigate other metrics to understand different aspects of the model. 

As examples, we also report two diversity scores: the proportion of distinct unigrams (\textbf{Dist-1}) and bigrams (\textbf{Dist-2}) in the generated replies~\citep{li2016diversity}.
While \abr{rouge} measures the relevance of the replies, higher diversity can also increase \abr{ctr}~\citep{deb2019diversifying}.
We can improve the diversity of the three replies with diversity-promoting decoding~\citep{li2016diversity,vijayakumar2018diverse,zhang2018generating} or latent variable models~\citep{deb2019diversifying}, but we leave this direction to future work.

\begin{figure*}[t]
\centering
\includegraphics[width=.7\textwidth]{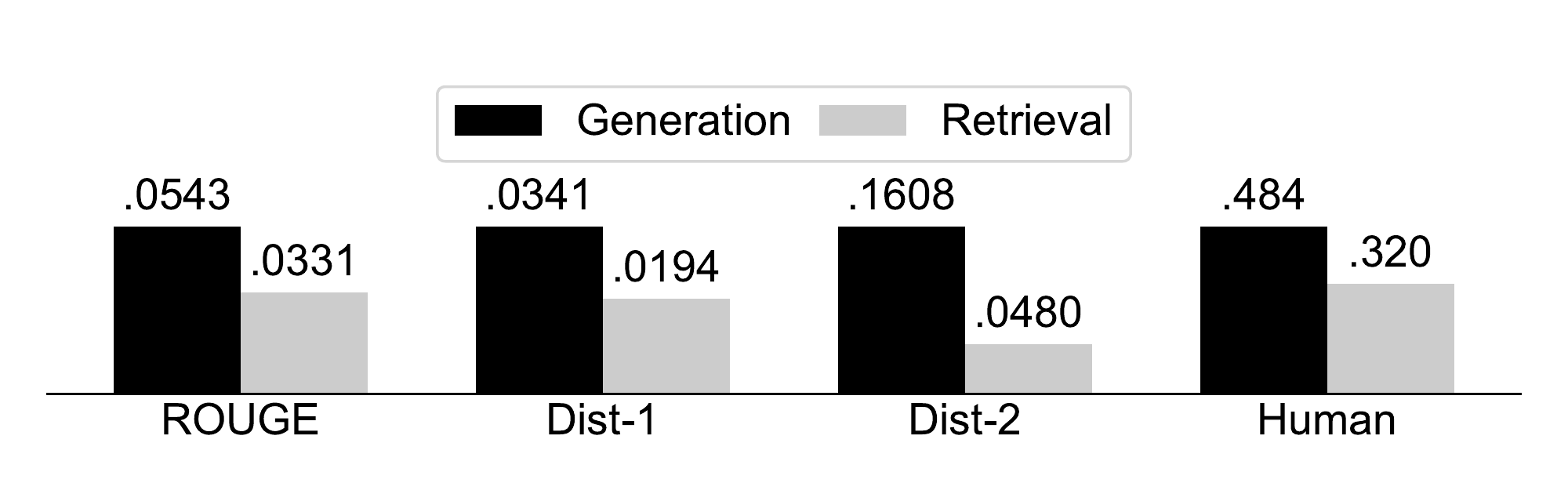}
\vspace{-1em}
\caption{Generation vs. retrieval model on English. Both automatic metrics and human referees prefer the generation model. The \textbf{human} score measures how often the human referee prefers the replies from one model over another.  The human score does not add up to one, because the referee can choose tie.}
\label{fig:en}
\end{figure*}

For our English monolingual experiments, we also complement automatic metrics with human judgments (\textbf{Human} in Figure~\ref{fig:en}). 
For each example, we display the input message and sets of three suggested replies from both generation and retrieval models to three human annotators (crowd workers).
We then ask the annotators to select the set with more responses that they prefer to send as a reply.
We leave evaluations for other languages to future work due to resource limitations.
\begin{table*}[t]
    \centering
    \setlength{\tabcolsep}{3pt}
    \begin{tabularx}{\textwidth}{lX}
    \toprule 
    \textbf{Message:} &  I think I want to play it so it better be available for PS4. \\
    \textbf{Generation:} & I'm sure it'll be available for PS4 as well. \\
    \textbf{Retrieval:} & I'd play it. \\
    \midrule
    \textbf{Message:} & Very nice! What model is this?  Obviously new Softail/Dyna, but which? \\
    \textbf{Generation:} & It's a Softail, but I'm not sure what model it is. \\
    \textbf{Retrieval:} & Delta. \\
    \midrule
    \textbf{Message:} & Interesting proposal. Especially since they just announced Mixon is going to have a scope. \\
    \textbf{Generation:} & I'm pretty sure it's going to have a scope, but I don't think it's going to have a scope.\\
    \textbf{Retrieval:} & Where did they say that? \\
    \midrule
    \textbf{Message:} & Oh the stickers! I envy you, Lush didn't sent me any with my body lotion order. \\
    \textbf{Generation:} & I'm so sorry. I'm so sorry. I'm so sorry. \\
    \textbf{Retrieval:} & What did you order? \\
    \bottomrule
    \end{tabularx}
    \caption{Example replies suggested by English models.  In the top two examples, the generation model produces more relevant replies than the retrieval model.  In the bottom two examples, the generation model produces contradictory and repetitive replies, while the retrieval model selects appropriate replies from the response set.}
    \label{tab:example}
\end{table*}

\begin{table*}[t]
    \centering
    \setlength{\tabcolsep}{3pt}
    \begin{tabular}{lrrrrrrrrrrrr}
        \toprule
        & \multicolumn{3}{c}{Monolingual} & \multicolumn{3}{c}{Zero-Shot} & \multicolumn{3}{c}{\abr{mt}} & \multicolumn{3}{c}{Multilingual}
        \\
        \cmidrule(lr){2-4}
        \cmidrule(lr){5-7}
        \cmidrule(lr){8-10}
        \cmidrule(lr){11-13}
        & \abr{rouge} & Dist-1 & Dist-2 & \abr{rouge} & Dist-1 & Dist-2 & \abr{rouge} & Dist-1 & Dist-2 & \abr{rouge} & Dist-1 & Dist-2 \\
        \midrule
        \flag{en} \abr{en} & \cellcolor{lightgrey} .0331 & \cellcolor{lightgrey} .0194 & \cellcolor{lightgrey} .0480 & \cellcolor{lightgrey} .0331 & \cellcolor{lightgrey} .0194 & \cellcolor{lightgrey} .0480 & - & - & - & \cellcolor{lightgrey} .0265 & \cellcolor{lightgrey} .0158 & \cellcolor{lightgrey} .0376 \\
        \flag{es} \abr{es} & \cellcolor{lightgrey} .0187 &  \cellcolor{lightgrey} .0157 &  \cellcolor{lightgrey} .0353 & \textbf{.0156} & .0113 & .0271 & .0139 & .0164 & .0350 & \cellcolor{lightgrey} .0181 & \cellcolor{lightgrey} .0151 & \cellcolor{lightgrey} .0333 \\
        \flag{de} \abr{de} & \cellcolor{lightgrey} .0215 & \cellcolor{lightgrey} .0134 & \cellcolor{lightgrey} .0298 & \textbf{.0178} & .0098 & .0240 & .0141 & .0152 & .0333 & \cellcolor{lightgrey} .0190 & \cellcolor{lightgrey} .0140 & \cellcolor{lightgrey} .0314 \\
        \flag{pt} \abr{pt} & \cellcolor{lightgrey} .0509 & \cellcolor{lightgrey} .0158 & \cellcolor{lightgrey} .0393 & \textbf{.0115} & .0121 & .0323 & .0110 & .0184 & .0449 & \cellcolor{lightgrey} .0460 & \cellcolor{lightgrey} .0161 & \cellcolor{lightgrey} .0401 \\
        \flag{fr} \abr{fr} & \cellcolor{lightgrey} .0216 & \cellcolor{lightgrey} .0191 & \cellcolor{lightgrey} .0468 & \textbf{.0168} & .0133 & .0343 & .0166 & .0196 & .0461 & \cellcolor{lightgrey} .0212 & \cellcolor{lightgrey} .0169 & \cellcolor{lightgrey} .0411 \\
        \flag{ja} \abr{ja} & \cellcolor{lightgrey} .0311 & \cellcolor{lightgrey} .0220 & \cellcolor{lightgrey} .0540 & \textbf{.0213} & .0236 & .0250 & .0153 & .1031 & .0444 & .0144 & .0677 & .0286 \\
        \flag{it} \abr{it} &  \cellcolor{lightgrey} .0200 &  \cellcolor{lightgrey} .0357 &  \cellcolor{lightgrey} .0768 & \textbf{.0172} & .0246 & .0576 & .0150 & .0378 & .0811 & .0171 & .0278 & .0614 \\
        \flag{sv} \abr{sv} &  \cellcolor{lightgrey} .0188 &  \cellcolor{lightgrey} .0287 &  \cellcolor{lightgrey} .0658 & .0168 & .0203 & .0506 & .0176 & .0302 & .0677 & \textbf{.0169} & .0224 & .0518 \\
        \flag{nl} \abr{nl} & \cellcolor{lightgrey} .0184 & \cellcolor{lightgrey} .0316 & \cellcolor{lightgrey} .0766 & .0167 & .0199 & .0533 & .0169 & .0297 & .0710 & \textbf{.0170} & .0221 & .0551 \\
        \flag{ru} \abr{ru} & \cellcolor{lightgrey} .0142 & \cellcolor{lightgrey} .0486 & \cellcolor{lightgrey} .0946 & .0138 & .0298 & .0604 & .0130 & .0431 & .0804 & \textbf{.0246} & .0405 & .0761 \\
        \bottomrule
    \end{tabular}
    \caption{Results for retrieval model initialized with \abr{mbert}~\citep{devlin2019bert}.
    The settings are in Section~\ref{ssec:setting}.
    \colorbox{lightgrey}{Gray cells} indicate when the model is trained on the target language training set.
    White cells indicate cross-lingual settings where the target language training set is not used for training.
    For each language, we \textbf{boldface} the best \abr{rouge} scores in cross-lingual settings (white cells).
    The zero-shot setting has better \abr{rouge} scores than using \abr{mt} data for most languages, and the results are sometimes close to monolingual training, confirming the effectiveness of \abr{mbert}.
    Multilingual training hurts training languages (gray cells compared to monolingual) but sometimes improves cross-lingual generalization (white cells compared to zero-shot).}
    \label{tab:ret}
\end{table*}
\begin{table*}[t]
    \centering
    \setlength{\tabcolsep}{3pt}
    \begin{tabular}{lrrrrrrrrr}
        \toprule
        & \multicolumn{3}{c}{Monolingual} & \multicolumn{3}{c}{\abr{mt}} & \multicolumn{3}{c}{Multilingual} \\
        \cmidrule(lr){2-4}
        \cmidrule(lr){5-7}
        \cmidrule(lr){8-10}
        & \abr{rouge} & \abr{dist1} & \abr{dist2} & \abr{rouge} & \abr{dist1} & \abr{dist2} & \abr{rouge} & \abr{dist1} & \abr{dist2} \\
        \midrule
        \flag{en} \abr{en} & \cellcolor{lightgrey} .0543 & \cellcolor{lightgrey} .0341 & \cellcolor{lightgrey} .161 & - & - & - & \cellcolor{lightgrey} .0412 & \cellcolor{lightgrey} .0352 & \cellcolor{lightgrey} .175 \\
        \flag{es} \abr{es} & \cellcolor{lightgrey} .0397 & \cellcolor{lightgrey} .0214 & \cellcolor{lightgrey} .182 & \textbf{.0270} & .0261 & .190 & \cellcolor{lightgrey} .0366 & \cellcolor{lightgrey} .0209 & \cellcolor{lightgrey} .175 \\
        \flag{de} \abr{de} & \cellcolor{lightgrey} .0469 & \cellcolor{lightgrey} .0332 & \cellcolor{lightgrey} .228 & \textbf{.0288} & .0244 & .142 & \cellcolor{lightgrey} .0454 & \cellcolor{lightgrey} .0321 & \cellcolor{lightgrey} .220 \\
        \flag{pt} \abr{pt} & \cellcolor{lightgrey} .0566 & \cellcolor{lightgrey} .0209 & \cellcolor{lightgrey} .194 & \textbf{.0276} & .0221 & .161 & \cellcolor{lightgrey} .0564 & \cellcolor{lightgrey} .0207 & \cellcolor{lightgrey} .190 \\
        \flag{fr} \abr{fr} & \cellcolor{lightgrey} .0446 & \cellcolor{lightgrey} .0207 & \cellcolor{lightgrey} .174 & \textbf{.0271} & .0165 & .109 & \cellcolor{lightgrey} .0428 & \cellcolor{lightgrey} .0211 & \cellcolor{lightgrey} .175 \\
        \flag{ja} \abr{ja} & \cellcolor{lightgrey} .0139 & \cellcolor{lightgrey} .1931 & \cellcolor{lightgrey} .245 & .0042 & .2812 & .216 & \textbf{.0114} & .0954 & .179 \\
        \flag{it} \abr{it} & \cellcolor{lightgrey} .0493 & \cellcolor{lightgrey} .0322 & \cellcolor{lightgrey} .243 & \textbf{.0316} & .0393 & .240 & .0295 & .0312 & .222 \\
        \flag{sv} \abr{sv} & \cellcolor{lightgrey} .0387 & \cellcolor{lightgrey} .0376 & \cellcolor{lightgrey} .236 & \textbf{.0369} & .0359 & .203 & .0241 & .0380 & .227 \\
        \flag{nl} \abr{nl} & \cellcolor{lightgrey} .0377 & \cellcolor{lightgrey} .0337 & \cellcolor{lightgrey} .230 & \textbf{.0320} & .0284 & .162 & .0233 & .0334 & .219 \\
        \flag{ru} \abr{ru} & \cellcolor{lightgrey} .0286 & \cellcolor{lightgrey} .0825 & \cellcolor{lightgrey} .349 & \textbf{.0238} & .0310 & .094 & .0165 & .0607 & .224 \\
        \bottomrule
    \end{tabular}
    \caption{Results for generation model.
    The settings are in Section~\ref{ssec:setting}.
    \colorbox{lightgrey}{Gray cells} indicate when the model is trained on the target language training set.
    White cells indicate cross-lingual settings where the target language training set is not used for training.
    For each language, we \textbf{boldface} the best \abr{rouge} scores in cross-lingual settings (white cells).
    Despite initializing with Unicoder-\abr{xdae}~\citep{liang2020xglue}, the model fails to generalize across languages in zero-shot settings.  The table does not include zero-shot results because the model only produces English replies and thus has near-zero \abr{rouge}.  Multilingual training hurts training languages (gray cells compared to monolingual), but the model can now generalize to unseen languages.  Training on \abr{mt} data is the best cross-lingual generalization method for the generation model.
    }
    \label{tab:gen}
\end{table*}

\section{Baseline Models}
\label{sec:model}

This section introduces the two baseline models: a retrieval model and a generation model.

\subsection{Retrieval Model}
\label{ssec:retrieval}

For the retrieval model, we use the architecture from \citet{henderson2017efficient}, except we replace the feedforward network encoders with Transformers~\citep{vaswani2017attention}.
Given an input message $\vect{x}$ and candidate reply $\vect{y}$, 
two Transformer encoders $\Phi_x$ and $\Phi_y$ map the message and the reply to two vectors $\Phi_x(\vect{x})$ and $\Phi_y(\vect{y})$.
The relevance score $\Theta_{\vect{x}\vect{y}}$ between the message $\vect{x}$ and the reply $\vect{y}$ is the dot product of the two vectors:
\begin{equation}
    \Theta_{\vect{x}\vect{y}} = \Phi_x(\vect{x})^\top \Phi_y(\vect{y}).
\end{equation}
\citet{henderson2017efficient} also adds a language model score to encourage more frequent replies.
We do not use language model score for simplicity.

We train the model with the symmetric loss from~\citet{deb2019diversifying}.
Suppose the batch size is $n$.
For a batch of training messages $\{\vect{x}_i\}_{i=1}^n$ and corresponding replies $\{\vect{y}_i\}_{j=1}^n$, we maximize:
\begin{equation}
    \sum_{i=1}^n \frac{e^{\Theta_{\vect{x}_i\vect{y}_i}}}{\sum_{j=1}^n \left (e^{\Theta_{\vect{x}_i\vect{y}_j}} + e^{\Theta_{\vect{x}_j\vect{y}_i}} \right) - e^{\Theta_{\vect{x}_i\vect{y}_i}}}.
\end{equation}
In a regular softmax loss, the denominator only sums over one variable.
The denominator in the symmetric loss sum over both variables to encourage bidirectional compatibility: the message should be predictive of the reply, and the reply should be predictive of the message.
This encourages the model to select responses specific to the message, similar to the Maximum Mutual Information objective from \citet{li2016diversity}.

The two encoders $\Phi_x$ and $\Phi_y$ are initialized with \abr{mbert}~\citep{devlin2019bert}, a Transformer with 110 million parameters pre-trained on multilingual corpora.
Initializing with \abr{mbert} allows the model to generalize across languages~\citep{wu2019beto}.
In Appendix~\ref{sec:xlmr}, we experiment with another pre-trained multilingual Transformer, \abr{xlm-r}~\citep{conneau2020unsupervised}.
We use the ``base'' version with 270 million parameters.

\subsection{Generation Model}

For the generation model, we follow the \abr{seq2seq} architecture (Section~\ref{ssec:framework}). 
We use a Transformer encoder to read the input $\vect{x}$, and another Transformer decoder to estimate $p(y_i \g \vect{x}, y_{<i})$ in \eqref{eq:likelihood}.

We cannot initialize the generation model with \abr{mbert} or \abr{xlm-r}, because the model also has a decoder.
Instead, we use Unicoder-\abr{xdae}~\citep{liang2020xglue}, a pre-trained multilingual \abr{seq2seq} model, which can generalize across languages in extractive generation tasks such as news title generation and question generation.
We test if Unicoder-\abr{xdae} also generalizes in the more challenging reply suggestion task.
There are other generation models we can use, which we discuss as future work in Section~\ref{sec:future}.

\subsection{Training Details}
\label{sec:hyperparameter}

We train the retrieval model using Adam optimizer~\citep{kingma2015adam} with 1e-6 learning rate, default $\beta$, and 256 batch size.
For monolingual and zero-shot settings, we use twenty epochs for English and fifty epochs for other languages.
We use ten epochs for \abr{mt} and multilingual settings.
The first 1\% training steps are warmup steps.
During training, we freeze the embedding layers and the bottom two Transformer layers of both encoders, which preserves multilingual knowledge from the pre-trained model and improves cross-lingual transfer learning~\citep{wu2019beto}.
All hyperparameters are manually tuned on the English validation set.

We use almost the same hyperparameters as \citet{liang2020xglue} to train generation models.
Specifically, we use Adam optimizer with 1e-5 initial learning rate, default $\beta$, and 1024 batch size.
For the monolingual and zero-shot setting, we use four epochs for English and 5000 steps for other languages (equivalent to two to nine epochs depending on the language).
We use one epoch for the \abr{mt} setting and 40,000 steps for the multilingual setting.
The first 20\% training steps are warmup steps.
We freeze the embedding layer during training for faster training.

All models are trained with eight Tesla V100 \abr{gpu}.
It takes about an hour to train the generation model for 1000 steps (covering about one million examples).
For the retrieval model, an epoch on the English training set (about 48 million examples) takes about seven hours.

\section{Results and Discussion}
\label{sec:result}

We experiment with the two baselines from Section~\ref{sec:model} on \name{}.
We first compare the models in English, where we have enough training data and human referees.
We then build models for other languages and compare training settings listed in Section~\ref{ssec:setting}.

\subsection{Results on English}
\label{ssec:en_exp}

Figure~\ref{fig:en} compares the generation and retrieval models in the English monolingual setting. 
Generation model not only has higher relevance (\abr{rouge}) score but also can generate more diverse replies (higher \abr{dist} scores).
For English, we also ask three human referees to compare the model outputs on a subset of 500 test examples.
Again, the referees prefer the generation model more often than the retrieval model (Figure~\ref{fig:en}).

We look at some generated responses to understand the models qualitatively.
In the top two examples in Table~\ref{tab:example}, the generation model produces replies highly specific to the input message.
In contrast, the retrieval model fails to find a relevant reply, because the response set does not cover these topics. 
This explains why the generation model has much higher \abr{rouge} and distinct $n$-gram scores than the retrieval model.

However, the expressiveness comes at the cost of a lack of control over the generated replies.
The generation model sometimes produces incoherent replies that are repetitive and/or contradictory, as shown in the bottom two examples of Table~\ref{tab:example}.
For the retrieval model, we can easily avoid these problems by curating the fixed response set.
These degenerative behaviors are observed in other text generation tasks and can be mitigated by changing training and decoding objectives~\citep{holtzman2020curious,welleck2019neural}.
We leave these directions for future research.

\subsection{Results on Other Languages}
\label{ssec:xling_exp}

After comparing English models, we experiment on other languages using the settings from Section~\ref{ssec:setting}.

\paragraph{Retrieval Model.}
Table~\ref{tab:ret} shows results for the retrieval model when initialized with \abr{mbert}.
The retrieval model can generalize fairly well across languages, as the \abr{rouge} in the zero-shot setting is often close to the monolingual setting.
This result confirms that initializing with \abr{mbert} is an effective strategy for cross-lingual generalization.
Training on \abr{mt} data is usually worse than training in the zero-shot setting.
This is possible because the \abr{mt} system may create artifacts that do not appear in organic data~\citep{artetxe2020translation}.
For the multilingual model, the training language \abr{rouge} scores are lower than monolingual training (gray cells in Table~\ref{tab:ret}).
However, multilingual training sometimes leads to better \abr{rouge} on unseen languages compared to transferring from only English (zero-shot).
Previous work observes similar results on other tasks, where multilingual training hurts training languages but helps generalization to unseen languages~\citep{johnson2017google,conneau2020unsupervised,wang2020negative}. 
Finally, Appendix~\ref{sec:xlmr} shows similar results when initializing with \abr{xlm-r}~\citep{conneau2020unsupervised}.

\paragraph{Generation Model.}
Table~\ref{tab:gen} shows results for the generation model.
In the monolingual setting, the generation model has higher scores than the retrieval model on most languages, consistent with the English result~(Figure~\ref{fig:en}).
However, unlike the retrieval model, the generation model fails to generalize across languages in the zero-shot setting, despite using Unicoder-\abr{xdae} for initialization.
We do not show zero-shot results in Table~\ref{tab:gen}, because \abr{rouge} are close to zero for non-English languages.
After training on English data, the model always produces English replies, regardless of the input language; i.e., the generation model ``forgets'' multilingual knowledge acquired during pre-training~\citep{kirkpatrick2017overcoming}.
This result is surprising because Unicoder-\abr{xdae} works in the zero-shot setting for other generation tasks~\citep{liang2020xglue},
which suggests that reply suggestion poses unique challenges for cross-lingual transfer learning. 
Interestingly, the multilingual model can generalize to unseen languages;
perhaps training on multiple languages regularizes the model to produce replies in the input language.
Overall, the best method to generalize the generation model across languages is to use machine-translated data.
\section{Future Work}
\label{sec:future}

\name{} opens up opportunities for future research.
Our experiments use four training settings (Section~\ref{ssec:setting}), but there are many other settings to explore.
For example, we can use other combinations of training languages, which may work better for some target languages~\citep{ammar2016many,cotterell2017cross,ahmad2019difficulties,lin2019choosing,zhang2020caco}.
We are also interested in training on both organic data and \abr{mt} data; i.e., mixing the zero-shot and \abr{mt} setting.

We can also compare other models on \name{}.
For the English monolingual setting, we can initialize the generation model with state-of-the-art language models~\citep{radford2019language,brown2020language,zhang2020dialogpt}.
For cross-lingual settings, we can initialize the generation model with several recent pre-trained multilingual \abr{seq2seq} models~\citep{chi2020cross,chi2021infoxlm,liu2020multilingual,tran2020cross,lewis2020marge,xue2020mt5}.
For retrieval models, we can experiment with other multilingual encoders that use different pre-training tasks~\citep{artetxe2019massively,chidambaram2019learning,reimers2020making,feng2020language}.

Another idea is to combine the two models.
Given an input message, we first use a generation model to create a set of candidate replies.
We then use a retrieval model to compute relevance scores and rerank these candidates.
Reranking the output of a generation model helps other natural language processing tasks~\citep{shen2004discriminative,collins2005discriminative,ge2006discriminative},
and previous work uses a similar idea for chatbots~\citep{qiu2017alime}.

Our experiment shows that reply suggestion poses unique challenges for cross-lingual generalization, especially for the generation model.
Future work can study methods to improve cross-lingual generalization methods. 
Some examples include applying adversarial learning~\citep{chen2018adversarial,chen2019multi,huang2019cross}, using adapters~\citep{pfeiffer2020mad}, adaptive transfer~\citep{xia2021metaxl}, mixing pre-training and fine-tuning~\citep{phang2020english}, and bringing a human in the loop~\citep{yuan2020interactive}.
\section{Conclusion}
\label{sec:conclusion}

We present \name{}, a multilingual dataset for reply suggestion.
We compare a generation and a retrieval baseline on \name{}.
The two models have different strengths in the English monolingual setting and require different strategies to transfer across languages.
\name{} provides a benchmark for future research in both reply suggestion and cross-lingual transfer learning.

\section*{Ethical Considerations}

\paragraph{Data Collection.}
No human annotators are involved while creating \name{}.
The examples and response sets of \name{} come from publicly available Reddit dumps from Pushshift, which are used in more than a hundred peer-reviewed publications~\citep{baumgartner2020pushshift}.

\paragraph{Privacy.}
Examples in \name{} do not have the username and are from publicly available data.
Therefore, we do not anticipate any privacy issues.
In the pilot study (Section~\ref{ssec:metric}), we measure the correlation of user \abr{ctr} with different evaluation metrics.
To protect user privacy, we only collect aggregated statistics (\abr{ctr}) and use no other information.

\paragraph{Potential Biased and Toxic Content.}
Despite our best effort to filter toxic contents (Section~\ref{ssec:filter}), the dataset may not be perfectly cleansed and may have other biases that are typical in open forums~\citep{massanari2017gamergate,mohan2017impact}.
Users should be aware of these issues.
We will continue to improve the quality of the dataset.

\paragraph{Intended Use of \name{}.}
Because of the possible biases and inappropriateness in the data, \name{} should \emph{not} be directly used to build production systems (as mentioned in Section~\ref{ssec:filter}).
The main use of \name{} is to test cross-lingual generalization for text retrieval and generation models, and researchers should be aware of possible ethical issues of Reddit data before using \name{}.

\section*{Acknowledgement}

We appreciate the feedback from anonymous reviewers.
MZ is supported in part by the Office of the Director of National Intelligence (\abr{odni}), Intelligence Advanced Research Projects Activity (\abr{iarpa}), via the \abr{better} Program contract \#2019-19051600005. The views and conclusions contained herein are those of the authors and should not be interpreted as necessarily representing the official policies, either expressed or implied, of \abr{odni}, \abr{iarpa}, or the U.S. Government. The U.S. Government is authorized to reproduce and distribute reprints for governmental purposes notwithstanding any copyright annotation therein.

\bibliographystyle{style/acl_natbib}
\bibliography{bib/journal-full,bib/mozhi}

\begin{thebibliography}{73}
\expandafter\ifx\csname natexlab\endcsname\relax\def\natexlab#1{#1}\fi

\bibitem[{Adiwardana et~al.(2020)Adiwardana, Luong, So, Hall, Fiedel,
  Thoppilan, Yang, Kulshreshtha, Nemade, Lu, and Le}]{adiwardana2020humanlike}
Daniel Adiwardana, Minh-Thang Luong, David~R. So, Jamie Hall, Noah Fiedel,
  Romal Thoppilan, Zi~Yang, Apoorv Kulshreshtha, Gaurav Nemade, Yifeng Lu, and
  Quoc~V. Le. 2020.
\newblock Towards a human-like open-domain chatbot.
\newblock \emph{arXiv preprint arXiv:2001.09977}.

\bibitem[{Ahmad et~al.(2019)Ahmad, Zhang, Ma, Hovy, Chang, and
  Peng}]{ahmad2019difficulties}
Wasi Ahmad, Zhisong Zhang, Xuezhe Ma, Eduard Hovy, Kai-Wei Chang, and Nanyun
  Peng. 2019.
\newblock \href {https://doi.org/10.18653/v1/N19-1253} {On difficulties of
  cross-lingual transfer with order differences: A case study on dependency
  parsing}.
\newblock In \emph{Conference of the North American Chapter of the Association
  for Computational Linguistics}.

\bibitem[{Ammar et~al.(2016)Ammar, Mulcaire, Ballesteros, Dyer, and
  Smith}]{ammar2016many}
Waleed Ammar, George Mulcaire, Miguel Ballesteros, Chris Dyer, and Noah~A.
  Smith. 2016.
\newblock \href {https://doi.org/10.1162/tacl_a_00109} {Many languages, one
  parser}.
\newblock \emph{Transactions of the Association for Computational Linguistics},
  4:431--444.

\bibitem[{Artetxe et~al.(2020)Artetxe, Labaka, and
  Agirre}]{artetxe2020translation}
Mikel Artetxe, Gorka Labaka, and Eneko Agirre. 2020.
\newblock \href {https://doi.org/10.18653/v1/2020.emnlp-main.618} {Translation
  artifacts in cross-lingual transfer learning}.
\newblock In \emph{Proceedings of Empirical Methods in Natural Language
  Processing}. Association for Computational Linguistics.

\bibitem[{Artetxe and Schwenk(2019)}]{artetxe2019massively}
Mikel Artetxe and Holger Schwenk. 2019.
\newblock \href {https://doi.org/10.1162/tacl_a_00288} {Massively multilingual
  sentence embeddings for zero-shot cross-lingual transfer and beyond}.
\newblock \emph{Transactions of the Association for Computational Linguistics},
  7:597--610.

\bibitem[{Banea et~al.(2008)Banea, Mihalcea, Wiebe, and
  Hassan}]{banea2008multilingual}
Carmen Banea, Rada Mihalcea, Janyce Wiebe, and Samer Hassan. 2008.
\newblock \href {https://www.aclweb.org/anthology/D08-1014} {Multilingual
  subjectivity analysis using machine translation}.
\newblock In \emph{Proceedings of Empirical Methods in Natural Language
  Processing}.

\bibitem[{Baumgartner et~al.(2020)Baumgartner, Zannettou, Keegan, Squire, and
  Blackburn}]{baumgartner2020pushshift}
Jason Baumgartner, Savvas Zannettou, Brian Keegan, Megan Squire, and Jeremy
  Blackburn. 2020.
\newblock The {P}ushshift {R}eddit dataset.
\newblock In \emph{International Conference on Weblogs and Social Media}.

\bibitem[{Brown et~al.(2020)Brown, Mann, Ryder, Subbiah, Kaplan, Dhariwal,
  Neelakantan, Shyam, Sastry, Askell et~al.}]{brown2020language}
Tom~B Brown, Benjamin Mann, Nick Ryder, Melanie Subbiah, Jared Kaplan, Prafulla
  Dhariwal, Arvind Neelakantan, Pranav Shyam, Girish Sastry, Amanda Askell,
  et~al. 2020.
\newblock Language models are few-shot learners.
\newblock \emph{arXiv preprint arXiv:2005.14165}.

\bibitem[{Chen et~al.(2019)Chen, Awadallah, Hassan, Wang, and
  Cardie}]{chen2019multi}
Xilun Chen, Ahmed~Hassan Awadallah, Hany Hassan, Wei Wang, and Claire Cardie.
  2019.
\newblock \href {https://doi.org/10.18653/v1/P19-1299} {Multi-source
  cross-lingual model transfer: Learning what to share}.
\newblock In \emph{Proceedings of the Association for Computational
  Linguistics}.

\bibitem[{Chen et~al.(2018)Chen, Sun, Athiwaratkun, Cardie, and
  Weinberger}]{chen2018adversarial}
Xilun Chen, Yu~Sun, Ben Athiwaratkun, Claire Cardie, and Kilian Weinberger.
  2018.
\newblock \href {https://doi.org/10.1162/tacl_a_00039} {Adversarial deep
  averaging networks for cross-lingual sentiment classification}.
\newblock \emph{Transactions of the Association for Computational Linguistics},
  6:557--570.

\bibitem[{Chi et~al.(2020)Chi, Dong, Wei, Wang, Mao, and Huang}]{chi2020cross}
Zewen Chi, Li~Dong, Furu Wei, Wenhui Wang, Xian-Ling Mao, and Heyan Huang.
  2020.
\newblock Cross-lingual natural language generation via pre-training.
\newblock In \emph{Association for the Advancement of Artificial Intelligence}.

\bibitem[{Chi et~al.(2021)Chi, Dong, Wei, Yang, Singhal, Wang, Song, Mao,
  Huang, and Zhou}]{chi2021infoxlm}
Zewen Chi, Li~Dong, Furu Wei, Nan Yang, Saksham Singhal, Wenhui Wang, Xia Song,
  Xian-Ling Mao, Heyan Huang, and Ming Zhou. 2021.
\newblock \href {https://www.aclweb.org/anthology/2021.naacl-main.280}
  {{I}nfo{XLM}: An information-theoretic framework for cross-lingual language
  model pre-training}.
\newblock In \emph{Conference of the North American Chapter of the Association
  for Computational Linguistics}.

\bibitem[{Chidambaram et~al.(2019)Chidambaram, Yang, Cer, Yuan, Sung, Strope,
  and Kurzweil}]{chidambaram2019learning}
Muthu Chidambaram, Yinfei Yang, Daniel Cer, Steve Yuan, Yunhsuan Sung, Brian
  Strope, and Ray Kurzweil. 2019.
\newblock \href {https://doi.org/10.18653/v1/W19-4330} {Learning cross-lingual
  sentence representations via a multi-task dual-encoder model}.
\newblock In \emph{Proceedings of ACL Workshop on Representation Learning for
  NLP}.

\bibitem[{Clark et~al.(2020)Clark, Choi, Collins, Garrette, Kwiatkowski,
  Nikolaev, and Palomaki}]{clark2020tydi}
Jonathan~H. Clark, Eunsol Choi, Michael Collins, Dan Garrette, Tom Kwiatkowski,
  Vitaly Nikolaev, and Jennimaria Palomaki. 2020.
\newblock \href {https://doi.org/10.1162/tacl_a_00317} {{T}y{D}i {QA}: A
  benchmark for information-seeking question answering in typologically diverse
  languages}.
\newblock \emph{Transactions of the Association for Computational Linguistics},
  8:454--470.

\bibitem[{Collins and Koo(2005)}]{collins2005discriminative}
Michael Collins and Terry Koo. 2005.
\newblock \href {https://doi.org/10.1162/0891201053630273} {Discriminative
  reranking for natural language parsing}.
\newblock \emph{Computational Linguistics}, 31(1):25--70.

\bibitem[{Conneau et~al.(2020)Conneau, Khandelwal, Goyal, Chaudhary, Wenzek,
  Guzm{\'a}n, Grave, Ott, Zettlemoyer, and Stoyanov}]{conneau2020unsupervised}
Alexis Conneau, Kartikay Khandelwal, Naman Goyal, Vishrav Chaudhary, Guillaume
  Wenzek, Francisco Guzm{\'a}n, Edouard Grave, Myle Ott, Luke Zettlemoyer, and
  Veselin Stoyanov. 2020.
\newblock \href {https://doi.org/10.18653/v1/2020.acl-main.747} {Unsupervised
  cross-lingual representation learning at scale}.
\newblock In \emph{Proceedings of the Association for Computational
  Linguistics}.

\bibitem[{Conneau et~al.(2018)Conneau, Lample, Rinott, Williams, Bowman,
  Schwenk, and Stoyanov}]{conneau2018xnli}
Alexis Conneau, Guillaume Lample, Ruty Rinott, Adina Williams, Samuel~R Bowman,
  Holger Schwenk, and Veselin Stoyanov. 2018.
\newblock \href {https://doi.org/10.18653/v1/D18-1269} {{XNLI}: Evaluating
  cross-lingual sentence representations}.
\newblock In \emph{Proceedings of Empirical Methods in Natural Language
  Processing}.

\bibitem[{Cotterell and Heigold(2017)}]{cotterell2017cross}
Ryan Cotterell and Georg Heigold. 2017.
\newblock \href {https://doi.org/10.18653/v1/D17-1078} {Cross-lingual
  character-level neural morphological tagging}.
\newblock In \emph{Proceedings of Empirical Methods in Natural Language
  Processing}.

\bibitem[{Deb et~al.(2019)Deb, Bailey, and Shokouhi}]{deb2019diversifying}
Budhaditya Deb, Peter Bailey, and Milad Shokouhi. 2019.
\newblock \href {https://doi.org/10.18653/v1/N19-2006} {Diversifying reply
  suggestions using a matching-conditional variational autoencoder}.
\newblock In \emph{Conference of the North American Chapter of the Association
  for Computational Linguistics (Industry Papers)}.

\bibitem[{Devlin et~al.(2019)Devlin, Chang, Lee, and
  Toutanova}]{devlin2019bert}
Jacob Devlin, Ming-Wei Chang, Kenton Lee, and Kristina Toutanova. 2019.
\newblock \href {https://doi.org/10.18653/v1/N19-1423} {{BERT}: Pre-training of
  deep bidirectional transformers for language understanding}.
\newblock In \emph{Conference of the North American Chapter of the Association
  for Computational Linguistics}. Association for Computational Linguistics.

\bibitem[{Dixon et~al.(2018)Dixon, Li, Sorensen, Thain, and
  Vasserman}]{dixon2018measuring}
Lucas Dixon, John Li, Jeffrey Sorensen, Nithum Thain, and Lucy Vasserman. 2018.
\newblock Measuring and mitigating unintended bias in text classification.
\newblock In \emph{Proceedings of AAAI/ACM Conference on AI, Ethics, and
  Society}.

\bibitem[{Feng et~al.(2020)Feng, Yang, Cer, Arivazhagan, and
  Wang}]{feng2020language}
Fangxiaoyu Feng, Yinfei Yang, Daniel Cer, Naveen Arivazhagan, and Wei Wang.
  2020.
\newblock Language-agnostic {BERT} sentence embedding.
\newblock \emph{arXiv preprint arXiv:2007.01852}.

\bibitem[{Gao et~al.(2019)Gao, Galley, Li et~al.}]{gao2019neural}
Jianfeng Gao, Michel Galley, Lihong Li, et~al. 2019.
\newblock Neural approaches to conversational {AI}.
\newblock \emph{Foundations and Trends in Information Retrieval},
  13(2-3):127--298.

\bibitem[{Ge and Mooney(2006)}]{ge2006discriminative}
Ruifang Ge and Raymond~J. Mooney. 2006.
\newblock \href {https://www.aclweb.org/anthology/P06-2034} {Discriminative
  reranking for semantic parsing}.
\newblock In \emph{Proceedings of the Association for Computational
  Linguistics}.

\bibitem[{Gehman et~al.(2020)Gehman, Gururangan, Sap, Choi, and
  Smith}]{gehman2020real}
Samuel Gehman, Suchin Gururangan, Maarten Sap, Yejin Choi, and Noah~A. Smith.
  2020.
\newblock \href {https://doi.org/10.18653/v1/2020.findings-emnlp.301}
  {{R}eal{T}oxicity{P}rompts: Evaluating neural toxic degeneration in language
  models}.
\newblock In \emph{Findings of the Association for Computational Linguistics:
  EMNLP}. Association for Computational Linguistics.

\bibitem[{Henderson et~al.(2017)Henderson, Al-Rfou, Strope, Sung, Luk{\'a}cs,
  Guo, Kumar, Miklos, and Kurzweil}]{henderson2017efficient}
Matthew Henderson, Rami Al-Rfou, Brian Strope, Yun-Hsuan Sung, L{\'a}szl{\'o}
  Luk{\'a}cs, Ruiqi Guo, Sanjiv Kumar, Balint Miklos, and Ray Kurzweil. 2017.
\newblock Efficient natural language response suggestion for {S}mart {R}eply.
\newblock \emph{arXiv preprint arXiv:1705.00652}.

\bibitem[{Holtzman et~al.(2020)Holtzman, Buys, Du, Forbes, and
  Choi}]{holtzman2020curious}
Ari Holtzman, Jan Buys, Li~Du, Maxwell Forbes, and Yejin Choi. 2020.
\newblock The curious case of neural text degeneration.
\newblock In \emph{Proceedings of the International Conference on Learning
  Representations}.

\bibitem[{Hu et~al.(2020)Hu, Ruder, Siddhant, Neubig, Firat, and
  Johnson}]{hu2020xtreme}
Junjie Hu, Sebastian Ruder, Aditya Siddhant, Graham Neubig, Orhan Firat, and
  Melvin Johnson. 2020.
\newblock {XTREME}: A massively multilingual multi-task benchmark for
  evaluating cross-lingual generalization.
\newblock In \emph{Proceedings of the International Conference of Machine
  Learning}.

\bibitem[{Huang et~al.(2019)Huang, Ji, and May}]{huang2019cross}
Lifu Huang, Heng Ji, and Jonathan May. 2019.
\newblock \href {https://doi.org/10.18653/v1/N19-1383} {Cross-lingual
  multi-level adversarial transfer to enhance low-resource name tagging}.
\newblock In \emph{Conference of the North American Chapter of the Association
  for Computational Linguistics}.

\bibitem[{Huang et~al.(2020)Huang, Zhu, and Gao}]{huang2020challenges}
Minlie Huang, Xiaoyan Zhu, and Jianfeng Gao. 2020.
\newblock Challenges in building intelligent open-domain dialog systems.
\newblock \emph{ACM Transactions on Information Systems}, 38(3):1--32.

\bibitem[{Hutchinson et~al.(2020)Hutchinson, Prabhakaran, Denton, Webster,
  Zhong, and Denuyl}]{hutchinson2020social}
Ben Hutchinson, Vinodkumar Prabhakaran, Emily Denton, Kellie Webster, Yu~Zhong,
  and Stephen Denuyl. 2020.
\newblock \href {https://doi.org/10.18653/v1/2020.acl-main.487} {Social biases
  in {NLP} models as barriers for persons with disabilities}.
\newblock In \emph{Proceedings of the Association for Computational
  Linguistics}.

\bibitem[{Johnson et~al.(2017)Johnson, Schuster, Le, Krikun, Wu, Chen, Thorat,
  Vi{\'e}gas, Wattenberg, Corrado, Hughes, and Dean}]{johnson2017google}
Melvin Johnson, Mike Schuster, Quoc~V. Le, Maxim Krikun, Yonghui Wu, Zhifeng
  Chen, Nikhil Thorat, Fernanda Vi{\'e}gas, Martin Wattenberg, Greg Corrado,
  Macduff Hughes, and Jeffrey Dean. 2017.
\newblock \href {https://doi.org/10.1162/tacl_a_00065} {{G}oogle{'}s
  multilingual neural machine translation system: Enabling zero-shot
  translation}.
\newblock \emph{Transactions of the Association for Computational Linguistics},
  5:339--351.

\bibitem[{Joulin et~al.(2016)Joulin, Grave, Bojanowski, Douze, J{\'e}gou, and
  Mikolov}]{joulin2016fasttext}
Armand Joulin, Edouard Grave, Piotr Bojanowski, Matthijs Douze, H{\'e}rve
  J{\'e}gou, and Tomas Mikolov. 2016.
\newblock {FastText.zip}: Compressing text classification models.
\newblock \emph{arXiv preprint arXiv:1612.03651}.

\bibitem[{Kannan et~al.(2016)Kannan, Kurach, Ravi, Kaufmann, Tomkins, Miklos,
  Corrado, Lukacs, Ganea, Young et~al.}]{kannan2016smart}
Anjuli Kannan, Karol Kurach, Sujith Ravi, Tobias Kaufmann, Andrew Tomkins,
  Balint Miklos, Greg Corrado, Laszlo Lukacs, Marina Ganea, Peter Young, et~al.
  2016.
\newblock Smart {R}eply: Automated response suggestion for email.
\newblock In \emph{Knowledge Discovery and Data Mining}.

\bibitem[{Kingma and Ba(2015)}]{kingma2015adam}
Diederik~P Kingma and Jimmy Ba. 2015.
\newblock Adam: A method for stochastic optimization.
\newblock In \emph{Proceedings of the International Conference on Learning
  Representations}.

\bibitem[{Kirkpatrick et~al.(2017)Kirkpatrick, Pascanu, Rabinowitz, Veness,
  Desjardins, Rusu, Milan, Quan, Ramalho, Grabska-Barwinska
  et~al.}]{kirkpatrick2017overcoming}
James Kirkpatrick, Razvan Pascanu, Neil Rabinowitz, Joel Veness, Guillaume
  Desjardins, Andrei~A Rusu, Kieran Milan, John Quan, Tiago Ramalho, Agnieszka
  Grabska-Barwinska, et~al. 2017.
\newblock Overcoming catastrophic forgetting in neural networks.
\newblock \emph{Proceedings of the National Academy of Sciences},
  114(13):3521--3526.

\bibitem[{Lewis et~al.(2020{\natexlab{a}})Lewis, Ghazvininejad, Ghosh,
  Aghajanyan, Wang, and Zettlemoyer}]{lewis2020marge}
Mike Lewis, Marjan Ghazvininejad, Gargi Ghosh, Armen Aghajanyan, Sida Wang, and
  Luke Zettlemoyer. 2020{\natexlab{a}}.
\newblock Pre-training via paraphrasing.
\newblock In \emph{Proceedings of Advances in Neural Information Processing
  Systems}.

\bibitem[{Lewis et~al.(2020{\natexlab{b}})Lewis, Oguz, Rinott, Riedel, and
  Schwenk}]{lewis2020mlqa}
Patrick Lewis, Barlas Oguz, Ruty Rinott, Sebastian Riedel, and Holger Schwenk.
  2020{\natexlab{b}}.
\newblock \href {https://doi.org/10.18653/v1/2020.acl-main.653} {{MLQA}:
  Evaluating cross-lingual extractive question answering}.
\newblock In \emph{Proceedings of the Association for Computational
  Linguistics}.

\bibitem[{Li et~al.(2016)Li, Galley, Brockett, Gao, and
  Dolan}]{li2016diversity}
Jiwei Li, Michel Galley, Chris Brockett, Jianfeng Gao, and Bill Dolan. 2016.
\newblock \href {https://doi.org/10.18653/v1/N16-1014} {A diversity-promoting
  objective function for neural conversation models}.
\newblock In \emph{Conference of the North American Chapter of the Association
  for Computational Linguistics}.

\bibitem[{Liang et~al.(2020)Liang, Duan, Gong, Wu, Guo, Qi, Gong, Shou, Jiang,
  Cao, Fan, Zhang, Agrawal, Cui, Wei, Bharti, Qiao, Chen, Wu, Liu, Yang,
  Campos, Majumder, and Zhou}]{liang2020xglue}
Yaobo Liang, Nan Duan, Yeyun Gong, Ning Wu, Fenfei Guo, Weizhen Qi, Ming Gong,
  Linjun Shou, Daxin Jiang, Guihong Cao, Xiaodong Fan, Ruofei Zhang, Rahul
  Agrawal, Edward Cui, Sining Wei, Taroon Bharti, Ying Qiao, Jiun-Hung Chen,
  Winnie Wu, Shuguang Liu, Fan Yang, Daniel Campos, Rangan Majumder, and Ming
  Zhou. 2020.
\newblock \href {https://doi.org/10.18653/v1/2020.emnlp-main.484} {{XGLUE}: A
  new benchmark datasetfor cross-lingual pre-training, understanding and
  generation}.
\newblock In \emph{Proceedings of Empirical Methods in Natural Language
  Processing}.

\bibitem[{Lin(2004)}]{lin2004rouge}
Chin-Yew Lin. 2004.
\newblock \href {https://www.aclweb.org/anthology/W04-1013} {{ROUGE}: A package
  for automatic evaluation of summaries}.
\newblock In \emph{Text Summarization Branches Out}.

\bibitem[{Lin et~al.(2019)Lin, Chen, Lee, Li, Zhang, Xia, Rijhwani, He, Zhang,
  Ma, Anastasopoulos, Littell, and Neubig}]{lin2019choosing}
Yu-Hsiang Lin, Chian-Yu Chen, Jean Lee, Zirui Li, Yuyan Zhang, Mengzhou Xia,
  Shruti Rijhwani, Junxian He, Zhisong Zhang, Xuezhe Ma, Antonios
  Anastasopoulos, Patrick Littell, and Graham Neubig. 2019.
\newblock \href {https://doi.org/10.18653/v1/P19-1301} {Choosing transfer
  languages for cross-lingual learning}.
\newblock In \emph{Proceedings of the Association for Computational
  Linguistics}.

\bibitem[{Liu et~al.(2016)Liu, Lowe, Serban, Noseworthy, Charlin, and
  Pineau}]{liu2016evaluate}
Chia-Wei Liu, Ryan Lowe, Iulian Serban, Mike Noseworthy, Laurent Charlin, and
  Joelle Pineau. 2016.
\newblock \href {https://doi.org/10.18653/v1/D16-1230} {How {NOT} to evaluate
  your dialogue system: An empirical study of unsupervised evaluation metrics
  for dialogue response generation}.
\newblock In \emph{Proceedings of Empirical Methods in Natural Language
  Processing}.

\bibitem[{Liu et~al.(2020)Liu, Gu, Goyal, Li, Edunov, Ghazvininejad, Lewis, and
  Zettlemoyer}]{liu2020multilingual}
Yinhan Liu, Jiatao Gu, Naman Goyal, Xian Li, Sergey Edunov, Marjan
  Ghazvininejad, Mike Lewis, and Luke Zettlemoyer. 2020.
\newblock \href {https://doi.org/10.1162/tacl_a_00343} {Multilingual denoising
  pre-training for neural machine translation}.
\newblock \emph{Transactions of the Association for Computational Linguistics},
  8:726--742.

\bibitem[{Massanari(2017)}]{massanari2017gamergate}
Adrienne Massanari. 2017.
\newblock \#{G}amergate and {T}he {F}appening: How {R}eddit’s algorithm,
  governance, and culture support toxic technocultures.
\newblock \emph{New Media \& Society}, 19(3):329--346.

\bibitem[{Mohan et~al.(2017)Mohan, Guha, Harris, Popowich, Schuster, and
  Priebe}]{mohan2017impact}
Shruthi Mohan, Apala Guha, Michael Harris, Fred Popowich, Ashley Schuster, and
  Chris Priebe. 2017.
\newblock The impact of toxic language on the health of {R}eddit communities.
\newblock In \emph{Canadian Conference on Artificial Intelligence}.

\bibitem[{Nivre et~al.(2016)Nivre, de~Marneffe, Ginter, Goldberg, Haji{\v{c}},
  Manning, McDonald, Petrov, Pyysalo, Silveira, Tsarfaty, and
  Zeman}]{nivre2016universal}
Joakim Nivre, Marie-Catherine de~Marneffe, Filip Ginter, Yoav Goldberg, Jan
  Haji{\v{c}}, Christopher~D. Manning, Ryan McDonald, Slav Petrov, Sampo
  Pyysalo, Natalia Silveira, Reut Tsarfaty, and Daniel Zeman. 2016.
\newblock \href {https://www.aclweb.org/anthology/L16-1262} {{U}niversal
  {D}ependencies v1: A multilingual treebank collection}.
\newblock In \emph{Proceedings of the Language Resources and Evaluation
  Conference}.

\bibitem[{Papineni et~al.(2002)Papineni, Roukos, Ward, and
  Zhu}]{papineni2002bleu}
Kishore Papineni, Salim Roukos, Todd Ward, and Wei-Jing Zhu. 2002.
\newblock \href {https://doi.org/10.3115/1073083.1073135} {{BLEU}: a method for
  automatic evaluation of machine translation}.
\newblock In \emph{Proceedings of the Association for Computational
  Linguistics}.

\bibitem[{Pfeiffer et~al.(2020)Pfeiffer, Vuli{\'c}, Gurevych, and
  Ruder}]{pfeiffer2020mad}
Jonas Pfeiffer, Ivan Vuli{\'c}, Iryna Gurevych, and Sebastian Ruder. 2020.
\newblock \href {https://doi.org/10.18653/v1/2020.emnlp-main.617} {{MAD-X}:
  {A}n {A}dapter-{B}ased {F}ramework for {M}ulti-{T}ask {C}ross-{L}ingual
  {T}ransfer}.
\newblock In \emph{Proceedings of Empirical Methods in Natural Language
  Processing}.

\bibitem[{Phang et~al.(2020)Phang, Calixto, Htut, Pruksachatkun, Liu, Vania,
  Kann, and Bowman}]{phang2020english}
Jason Phang, Iacer Calixto, Phu~Mon Htut, Yada Pruksachatkun, Haokun Liu, Clara
  Vania, Katharina Kann, and Samuel~R. Bowman. 2020.
\newblock \href {https://www.aclweb.org/anthology/2020.aacl-main.56} {{E}nglish
  intermediate-task training improves zero-shot cross-lingual transfer too}.
\newblock In \emph{Conference of the Asia-Pacific Chapter of the Association
  for Computational Linguistics}.

\bibitem[{Qiu et~al.(2017)Qiu, Li, Wang, Gao, Chen, Zhao, Chen, Huang, and
  Chu}]{qiu2017alime}
Minghui Qiu, Feng-Lin Li, Siyu Wang, Xing Gao, Yan Chen, Weipeng Zhao, Haiqing
  Chen, Jun Huang, and Wei Chu. 2017.
\newblock \href {https://doi.org/10.18653/v1/P17-2079} {{A}li{M}e chat: A
  sequence to sequence and rerank based chatbot engine}.
\newblock In \emph{Proceedings of the Association for Computational
  Linguistics}.

\bibitem[{Radford et~al.(2019)Radford, Wu, Child, Luan, Amodei, and
  Sutskever}]{radford2019language}
Alec Radford, Jeffrey Wu, Rewon Child, David Luan, Dario Amodei, and Ilya
  Sutskever. 2019.
\newblock Language models are unsupervised multitask learners.
\newblock \emph{OpenAI Blog}.

\bibitem[{Reimers and Gurevych(2020)}]{reimers2020making}
Nils Reimers and Iryna Gurevych. 2020.
\newblock \href {https://doi.org/10.18653/v1/2020.emnlp-main.365} {Making
  monolingual sentence embeddings multilingual using knowledge distillation}.
\newblock In \emph{Proceedings of Empirical Methods in Natural Language
  Processing}.

\bibitem[{Sap et~al.(2019)Sap, Card, Gabriel, Choi, and Smith}]{sap2019risk}
Maarten Sap, Dallas Card, Saadia Gabriel, Yejin Choi, and Noah~A. Smith. 2019.
\newblock \href {https://doi.org/10.18653/v1/P19-1163} {The risk of racial bias
  in hate speech detection}.
\newblock In \emph{Proceedings of the Association for Computational
  Linguistics}.

\bibitem[{Schwenk and Li(2018)}]{li2018mldoc}
Holger Schwenk and Xian Li. 2018.
\newblock \href {https://www.aclweb.org/anthology/L18-1560} {A corpus for
  multilingual document classification in eight languages}.
\newblock In \emph{Proceedings of the Language Resources and Evaluation
  Conference}.

\bibitem[{Scialom et~al.(2020)Scialom, Dray, Lamprier, Piwowarski, and
  Staiano}]{scialom2020mlsum}
Thomas Scialom, Paul-Alexis Dray, Sylvain Lamprier, Benjamin Piwowarski, and
  Jacopo Staiano. 2020.
\newblock \href {https://doi.org/10.18653/v1/2020.emnlp-main.647} {{MLSUM}: The
  multilingual summarization corpus}.
\newblock In \emph{Proceedings of Empirical Methods in Natural Language
  Processing}.

\bibitem[{Shen et~al.(2004)Shen, Sarkar, and Och}]{shen2004discriminative}
Libin Shen, Anoop Sarkar, and Franz~Josef Och. 2004.
\newblock \href {https://www.aclweb.org/anthology/N04-1023} {Discriminative
  reranking for machine translation}.
\newblock In \emph{Conference of the North American Chapter of the Association
  for Computational Linguistics}.

\bibitem[{Strassel and Tracey(2016)}]{strassel2016lorelei}
Stephanie Strassel and Jennifer Tracey. 2016.
\newblock \href {https://www.aclweb.org/anthology/L16-1521} {{LORELEI} language
  packs: Data, tools, and resources for technology development in low resource
  languages}.
\newblock In \emph{Proceedings of the Language Resources and Evaluation
  Conference}.

\bibitem[{Sutskever et~al.(2014)Sutskever, Vinyals, and
  Le}]{sutskever2014sequence}
Ilya Sutskever, Oriol Vinyals, and Quoc~V Le. 2014.
\newblock Sequence to sequence learning with neural networks.
\newblock In \emph{Proceedings of Advances in Neural Information Processing
  Systems}.

\bibitem[{Tjong Kim~Sang(2002)}]{tjong2002introduction}
Erik~F. Tjong Kim~Sang. 2002.
\newblock \href {https://www.aclweb.org/anthology/W02-2024} {Introduction to
  the {C}o{NLL}-2002 shared task: Language-independent named entity
  recognition}.
\newblock In \emph{Conference on Computational Natural Language Learning}.

\bibitem[{Tran et~al.(2020)Tran, Tang, Li, and Gu}]{tran2020cross}
Chau Tran, Yuqing Tang, Xian Li, and Jiatao Gu. 2020.
\newblock Cross-lingual retrieval for iterative self-supervised training.
\newblock In \emph{Proceedings of Advances in Neural Information Processing
  Systems}.

\bibitem[{Vaswani et~al.(2017)Vaswani, Shazeer, Parmar, Uszkoreit, Jones,
  Gomez, Kaiser, and Polosukhin}]{vaswani2017attention}
Ashish Vaswani, Noam Shazeer, Niki Parmar, Jakob Uszkoreit, Llion Jones,
  Aidan~N Gomez, {\L}ukasz Kaiser, and Illia Polosukhin. 2017.
\newblock Attention is all you need.
\newblock In \emph{Proceedings of Advances in Neural Information Processing
  Systems}.

\bibitem[{Vijayakumar et~al.(2018)Vijayakumar, Cogswell, Selvaraju, Sun, Lee,
  Crandall, and Batra}]{vijayakumar2018diverse}
Ashwin~K Vijayakumar, Michael Cogswell, Ramprasath~R Selvaraju, Qing Sun,
  Stefan Lee, David Crandall, and Dhruv Batra. 2018.
\newblock Diverse beam search: Decoding diverse solutions from neural sequence
  models.
\newblock In \emph{Association for the Advancement of Artificial Intelligence}.

\bibitem[{Wang et~al.(2020)Wang, Lipton, and Tsvetkov}]{wang2020negative}
Zirui Wang, Zachary~C. Lipton, and Yulia Tsvetkov. 2020.
\newblock \href {https://doi.org/10.18653/v1/2020.emnlp-main.359} {On negative
  interference in multilingual models: Findings and a meta-learning treatment}.
\newblock In \emph{Proceedings of Empirical Methods in Natural Language
  Processing}, Online. Association for Computational Linguistics.

\bibitem[{Welleck et~al.(2020)Welleck, Kulikov, Roller, Dinan, Cho, and
  Weston}]{welleck2019neural}
Sean Welleck, Ilia Kulikov, Stephen Roller, Emily Dinan, Kyunghyun Cho, and
  Jason Weston. 2020.
\newblock Neural text generation with unlikelihood training.
\newblock In \emph{Proceedings of the International Conference on Learning
  Representations}.

\bibitem[{Wu and Dredze(2019)}]{wu2019beto}
Shijie Wu and Mark Dredze. 2019.
\newblock \href {https://doi.org/10.18653/v1/D19-1077} {Beto, bentz, becas: The
  surprising cross-lingual effectiveness of {BERT}}.
\newblock In \emph{Proceedings of Empirical Methods in Natural Language
  Processing}.

\bibitem[{Xia et~al.(2021)Xia, Zheng, Mukherjee, Shokouhi, Newbig, and
  Awadallah}]{xia2021metaxl}
Mengzhou Xia, Guoqing Zheng, Subhabrata Mukherjee, Milad Shokouhi, Graham
  Newbig, and Ahmed~Hassan Awadallah. 2021.
\newblock \href {https://www.aclweb.org/anthology/2021.naacl-main.42}
  {Meta{XL}: Meta representation transformation for low-resource cross-lingual
  learning}.
\newblock In \emph{Conference of the North American Chapter of the Association
  for Computational Linguistics}.

\bibitem[{Xue et~al.(2020)Xue, Constant, Roberts, Kale, Al-Rfou, Siddhant,
  Barua, and Raffel}]{xue2020mt5}
Linting Xue, Noah Constant, Adam Roberts, Mihir Kale, Rami Al-Rfou, Aditya
  Siddhant, Aditya Barua, and Colin Raffel. 2020.
\newblock {mT5}: A massively multilingual pre-trained text-to-text transformer.
\newblock \emph{arXiv preprint arXiv:2010.11934}.

\bibitem[{Yuan et~al.(2020)Yuan, Zhang, Durme, Findlater, and
  Boyd-Graber}]{yuan2020interactive}
Michelle Yuan, Mozhi Zhang, Benjamin~Van Durme, Leah Findlater, and Jordan
  Boyd-Graber. 2020.
\newblock \href {https://doi.org/10.18653/v1/2020.emnlp-main.482} {Interactive
  refinement of cross-lingual word embeddings}.
\newblock In \emph{Proceedings of Empirical Methods in Natural Language
  Processing}.

\bibitem[{Zhang et~al.(2020{\natexlab{a}})Zhang, Fujinuma, and
  Boyd-Graber}]{zhang2020caco}
Mozhi Zhang, Yoshinari Fujinuma, and Jordan Boyd-Graber. 2020{\natexlab{a}}.
\newblock Exploiting cross-lingual subword similarities in low-resource
  document classification.
\newblock In \emph{Association for the Advancement of Artificial Intelligence}.

\bibitem[{Zhang et~al.(2020{\natexlab{b}})Zhang, Fujinuma, Paul, and
  Boyd-Graber}]{zhang2020overfit}
Mozhi Zhang, Yoshinari Fujinuma, Michael~J. Paul, and Jordan Boyd-Graber.
  2020{\natexlab{b}}.
\newblock \href {https://doi.org/10.18653/v1/2020.acl-main.201} {Why
  overfitting isn't always bad: Retrofitting cross-lingual word embeddings to
  dictionaries}.
\newblock In \emph{Proceedings of the Association for Computational
  Linguistics}.

\bibitem[{Zhang et~al.(2018)Zhang, Galley, Gao, Gan, Li, Brockett, and
  Dolan}]{zhang2018generating}
Yizhe Zhang, Michel Galley, Jianfeng Gao, Zhe Gan, Xiujun Li, Chris Brockett,
  and Bill Dolan. 2018.
\newblock Generating informative and diverse conversational responses via
  adversarial information maximization.
\newblock In \emph{Proceedings of Advances in Neural Information Processing
  Systems}.

\bibitem[{Zhang et~al.(2020{\natexlab{c}})Zhang, Sun, Galley, Chen, Brockett,
  Gao, Gao, Liu, and Dolan}]{zhang2020dialogpt}
Yizhe Zhang, Siqi Sun, Michel Galley, Yen-Chun Chen, Chris Brockett, Xiang Gao,
  Jianfeng Gao, Jingjing Liu, and Bill Dolan. 2020{\natexlab{c}}.
\newblock \href {https://doi.org/10.18653/v1/2020.acl-demos.30} {{D}ialo{GPT}:
  Large-scale generative pre-training for conversational response generation}.
\newblock In \emph{Proceedings of the Association for Computational
  Linguistics: System Demonstrations}.

\end{thebibliography}

\clearpage
\begin{appendix}
  \section{Results for \abr{xlm-r}}
\label{sec:xlmr}
\begin{minipage}{2\linewidth}
    \centering
    \setlength{\tabcolsep}{3pt}
    \begin{tabular}{lrrrrrrrrrrrr}
        \toprule
        & \multicolumn{3}{c}{Monolingual} & \multicolumn{3}{c}{Zero-Shot} & \multicolumn{3}{c}{\abr{mt}} & \multicolumn{3}{c}{Multilingual}
        \\
        \cmidrule(lr){2-4}
        \cmidrule(lr){5-7}
        \cmidrule(lr){8-10}
        \cmidrule(lr){11-13}
        & \abr{rouge} & Dist-1 & Dist-2 & \abr{rouge} & Dist-1 & Dist-2 & \abr{rouge} & Dist-1 & Dist-2 & \abr{rouge} & Dist-1 & Dist-2 \\
        \midrule
        \flag{en} \abr{en} & \cellcolor{lightgrey} .0354 & \cellcolor{lightgrey} .0177 & \cellcolor{lightgrey} .0454 & \cellcolor{lightgrey} .0354 & \cellcolor{lightgrey} .0177 & \cellcolor{lightgrey} .0454 & - & - & - & \cellcolor{lightgrey} .0319 & \cellcolor{lightgrey} .0152 & \cellcolor{lightgrey} .0398 \\
        \flag{es} \abr{es} & \cellcolor{lightgrey} .0158 & \cellcolor{lightgrey} .0069 & \cellcolor{lightgrey} .0172 & \textbf{.0140} & .0065 & .0160 & .0122 & .0079 & .0181 & \cellcolor{lightgrey} .0155 & \cellcolor{lightgrey} .0076 & \cellcolor{lightgrey} .0182 \\
        \flag{de} \abr{de} & \cellcolor{lightgrey} .0179 & \cellcolor{lightgrey} .0098 & \cellcolor{lightgrey} .0261 & \textbf{.0141} & .0064 & .0162 & .0132 & .0071 & .0170 & \cellcolor{lightgrey} .0171 & \cellcolor{lightgrey} .0069 & \cellcolor{lightgrey} .0170 \\
        \flag{pt} \abr{pt} & \cellcolor{lightgrey} .0345 & \cellcolor{lightgrey} .0088 & \cellcolor{lightgrey} .0239 & \textbf{.0126} & .0076 & .0209 & .0120 & .0071 & .0178 & \cellcolor{lightgrey} .0332 & \cellcolor{lightgrey} .0086 & \cellcolor{lightgrey} .0230 \\
        \flag{fr} \abr{fr} & \cellcolor{lightgrey} .0161 & \cellcolor{lightgrey} .0062 & \cellcolor{lightgrey} .0168 & \textbf{.0143} & .0066 & .0177 & .0135 & .0073 & .0184 & \cellcolor{lightgrey} .0161 & \cellcolor{lightgrey} .0069 & \cellcolor{lightgrey} .0185 \\
        \flag{ja} \abr{ja} & \cellcolor{lightgrey} .0271 & \cellcolor{lightgrey} .0132 & \cellcolor{lightgrey} .0364 & \textbf{.0181} & .0097 & .0277 & .0157 & .0106 & .0293 &.0166 & .0123 & .0328 \\
        \flag{it} \abr{it} & \cellcolor{lightgrey} .0157 & \cellcolor{lightgrey} .0123 & \cellcolor{lightgrey} .0291 & \textbf{.0144} & .0123 & .0306 & .0155 & .0156 & .0375 & .0143 & .0136 & .0337 \\
        \flag{sv} \abr{sv} & \cellcolor{lightgrey} .0172 & \cellcolor{lightgrey} .0129 & \cellcolor{lightgrey} .0333 & .0165 & .0133 & .0333 & .0153 & .0140 & .0341 & \textbf{.0168} & .0125 & .0321 \\
        \flag{nl} \abr{nl} & \cellcolor{lightgrey} .0171 & \cellcolor{lightgrey} .0142 & \cellcolor{lightgrey} .0390 & .0161 & .0134 & .0371 & .0155 & .0134 & .0353 & \textbf{.0162} & .0135 & .0370 \\
        \flag{ru} \abr{ru} & \cellcolor{lightgrey} .0128 & \cellcolor{lightgrey} .0259 & \cellcolor{lightgrey} .0541 & .0123 & .0223 & .0467 & .0111 & .0248 & .0506 & \textbf{.0130} & .0244 & .0510 \\
        \bottomrule
    \end{tabular}
    \captionof{table}{Results for retrieval model initialized with \abr{xlm-r}~\citep{conneau2020unsupervised},
    The settings are in Section~\ref{ssec:setting}.
    \colorbox{lightgrey}{Gray cells} indicate when the model is trained on the target language training set.
    White cells indicate cross-lingual settings where the target language training set is not used for training.
    For each language, we \textbf{boldface} the best \abr{rouge} scores in cross-lingual settings (white cells).
    We observe similar trends as \abr{mbert} (Table~\ref{tab:ret}).}
    \label{tab:xlmr}
\end{minipage}
\end{appendix}

\end{document}